\documentclass[fleqn,10pt]{olplainarticle}

\usepackage{tikz}
\usetikzlibrary{positioning}

\usepackage[utf8]{inputenc}
\usepackage[T1]{fontenc}

\usepackage{geometry}
\usepackage{color}
\usepackage{color,soul}
\usepackage{amsmath}

\usepackage{color}
\usepackage{graphicx}
\usepackage{amsfonts}
\usepackage{bm}
\usepackage{mathtools}
\usepackage{amsthm}

\usepackage{wrapfig}
\usepackage{hyperref}
\hypersetup{
    colorlinks=true,
    linkcolor=blue,
    filecolor=magenta,      
    urlcolor=cyan,
    pdftitle={Overleaf Example},
    pdfpagemode=FullScreen,
    }

\title{Evaluation of machine learning architectures on the quantification of epistemic and aleatoric uncertainties in complex dynamical systems}

\author[1]{Stephen Guth}
\author[1]{Alireza Mojahed}
\author[1]{Themistoklis P. Sapsis*}
\affil[1]{Department of Mechanical Engineering, Massachusetts Institute of Technology, Cambridge, MA 02139, USA}
\affil[*]{corresponding author, sapsis@mit.edu}


\begin{abstract}
Machine learning methods for the construction of data-driven reduced order model models are used in an increasing variety of engineering domains, especially as a supplement to expensive computational fluid dynamics for design problems.  An important check on the reliability of surrogate models is Uncertainty Quantification (UQ), a self assessed estimate of the model error.  Accurate UQ allows for cost savings by reducing both the required size of training data sets and the required safety factors, while poor UQ prevents users from confidently relying on model predictions.  We examine several machine learning techniques, including both Gaussian processes and a family UQ-augmented neural networks:  Ensemble neural networks (ENN), Bayesian neural networks (BNN), Dropout neural networks (D-NN), and Gaussian neural networks (G-NN).  We evaluate UQ accuracy (distinct from model accuracy) using two metrics:  the distribution of normalized residuals on validation data, and the distribution of estimated uncertainties.  We apply these metrics to two model data sets, representative of complex dynamical systems:  an ocean engineering problem in which a ship traverses irregular wave episodes, and a dispersive wave turbulence system with extreme events, the Majda-McLaughlin-Tabak model.  We present conclusions concerning model architecture and hyperparameter tuning.
\end{abstract}

\begin{document}

\flushbottom
\maketitle
\thispagestyle{empty}


\section{Introduction}

For a wide variety of engineering problems, numerical predictions are only a first step--also necessary is an estimate of the result error.  The economic costs of unconstrained errors are evidenced by the costs of conservative tolerances and safety factors.  For design problems, uncertainty quantification (UQ) is the general problem of estimating the magnitude of the error between model results and the (unobserved) ground truth \cite{berger19, abdar21}, and it has wide application across many modeling domains \cite{kennamer2019, wang19, akbari20, charalampopoulos22}.  Traditional design methodologies based on theory or simulation allow for error propagation, sensitivity analysis, and convergence studies, however these approaches are difficult to extend to the black box techniques currently available from machine learning.

For UQ, the applied machine learning literature has primarily drawn from Gaussian Process (GP) techniques \cite{rasmussen05}, which have built-in UQ in the form of a posterior distribution \cite{gong22, guth22, guth23}.  GP techniques, sometimes called kriging, are especially useful for active learning (optimal experimental design) problems \cite{mohamad18, blanchard20e, blanchard21, yang22} and multi-fidelity problems \cite{perdikaris15,babaee20,champenois23}.  However, GP techniques are limited in two significant ways:  the $O(n_s^3)$ scaling of computation time with the number of training samples, and the more subtle breakdown with the dimensionality of the problem input \cite{pickering22a}.

The other main approach in the machine learning literature is to augment neural networks with some kind of uncertainty quantification \cite{abdar21, psaros23}. The first thread is to use ensemble techniques to implement the Bayes Model Average from Variational Inference \cite{pickering22a, pickering22}.  The second thread is to combine GP and neural network technology to combine the scalability of neural networks with the posterior distribution of GP.  Various works in the direction include the marginal likelihood loss function \cite{tomczak21, schwobel22} the sparse inducing point framework \cite{hensman15, wilson15, burt20, ober21}, Deep Gaussian Process \cite{damianou13, bradshaw17}, and Deep Kernel Learning \cite{amersfoort21, ober21a}.  These techniques seek to avoid the $O(n_s^3)$ scaling costs of exact GP inference by replacing exact calculations with approximations. We also refer to recent studies \cite{psaros23, zou22} for a comprehensive comparison for various scientific ML data sets.

In this work, we compare the UQ skill of selected data-driven methods as applied to two datasets associated with complex dynamical systems from ocean engineering and turbulent systems. While previous works have also considered comparison of different UQ methods using NN architectures \cite{psaros23, zou22}, they primarily focus on low-complexity dynamical systems (e.g. KdV equation) or problems with very small uncertainty and without a strongly turbulent character. The chosen problems here are characterized by high complexity in the sense that the uncertainty of the input variables is  large (i.e. not of a perturbation type), while for the first data set there is also important aleatoric uncertainty, and for the second data set there are intermittent instabilities leading to highly non-linear extreme events. In this sense the data sets have a mix of uncertainty types (measurement noise vs noise free) and a range of dimensions (from $n=2$ to $n=20$).  We compare GP against three deep neural network -based  techniques for estimating epistemic uncertainty:  ensemble networks, Bayesian networks, and dropout networks; as well as one technique for estimating aleatoric uncertainty:  Gaussian networks.  
 
 In order to holistically evaluate the UQ, we introduce two criteria that characterize the UQ performance for these methods: i) the distribution of normalized residuals on validation data, which measures how accurately the model estimates its own errors, and ii) the distribution of epistemic uncertainties, which measures the typical magnitude of the model errors.

The structure of this paper is as follows.  First, in section \ref{sec:uq}, we present the typology of uncertainty we follow in this work, as well as the two datasets we use for evaluation.  Next, in section \ref{sec:ml-arch}, we describe each of the machine learning architectures we examine below.  In section \ref{sec:metrics}, we briefly describe the metrics we plot to examine the UQ calibration.  Finally, in section \ref{sec:results}, we present our results.  We additionally include an appendix with alternate visualizations of the uncertainty of each surrogate model in 2D restrictions of each dataset.

\section{Uncertainty Quantification}
\label{sec:uq}

\subsection{Typology of Uncertainty}

In a wide variety of modeling scenarios, it is not only important that the model be accurate (that is, generate predictions with low errors), but that the model provide accurate estimates for the its own errors.  While UQ, and accurate UQ, is important for many reasons, we choose three to highlight.  First, for applications that require high reliability, accurate UQ can help engineers to decide if the model reliability is sufficient for the task.  Second, for some statistical tasks, measuring the uncertainty of the model is as important or more important than measuring the mean prediction.  Third, an important strand of optimal experimental design involves choosing new experimental designs in regions of the parameter space where the model is most uncertain. While each of these tasks require UQ, they each examine a subtly different flavor of uncertainty.  In summarizing a long literature of describing uncertainty \cite{hullermeier21}, we distinguish between two types of uncertainty:  aleatoric uncertainty and epistemic uncertainty.

Aleatoric uncertainty, sometimes called data uncertainty, is the irreducible uncertainty associated with unmeasureable randomness.  The classic example is an experiment in coin flipping.  No matter how much data you collect, the uncertainty in your estimate of which face the coin will show after its next flip will never decrease.  In this paper, aleatoric uncertainty is the spread in outcomes we might expect to measure if we performed the same experiment more than once.


Epistemic uncertainty, sometimes called model uncertainty, is the uncertainty associated with lack of knowledge or sparse data.  For many types experiments, we expect that experiments nearby to existing samples in parameter space will usually yield similar results.  At the same time, as new experimental outputs may differ more and more from the existing data we become less and less certain what the results might be.  Epistemic uncertainty is exhibited when many possible models explain the data we already have, but offer different predictions for unseen data.  In general, we reduce epistemic uncertainty by collecting more data--particularly, data from novel experimental designs.


These two measures of uncertainty have different applications.  For active sampling procedures, the epistemic uncertainty is the natural quantity of interest--choose the experimental design that will teach us the most.  For statistical problems, the aleatoric uncertainty stands in the forefront.  For many steady state problems, the aleatoric uncertainty itself is the quantity we want to measure.  Finally, for reliability questions, we are interested in the total uncertainty--the sum of aleatoric and epistemic.  The careful typology of uncertainty is important \cite{berger19}, because different techniques may measure epistemic and aleatoric uncertainties differently.  In this paper, we generally refer to the standard deviation of the aleatoric uncertainty as $\sigma_n$ and the corresponding epistemic standard deviation as $\sigma_{\epsilon}$.

The UQ challenge for model design with machine learning techniques is to produce reasonable estimates of the epistemic and aleatoric uncertainty.  The standard neural network provides \textit{no} estimate of its own uncertainty;  this stands in contrast with GP models, which provide extensive estimates of their own uncertainties.  In this paper, we summarize the UQ from different machine learning techniques both on a reference ocean engineering problem, one where both epistemic and aleatoric uncertainty are significant, and a reference fluid dynamics problem, where aleatoric uncertainty is known to be negligible.

\subsection{Ship loads in irregular seas data set (LAMP data set)}
\label{sec:lamp-data}

\begin{figure}[]
\centering
\includegraphics[width=0.4\linewidth, trim=0 0 0 0, clip] {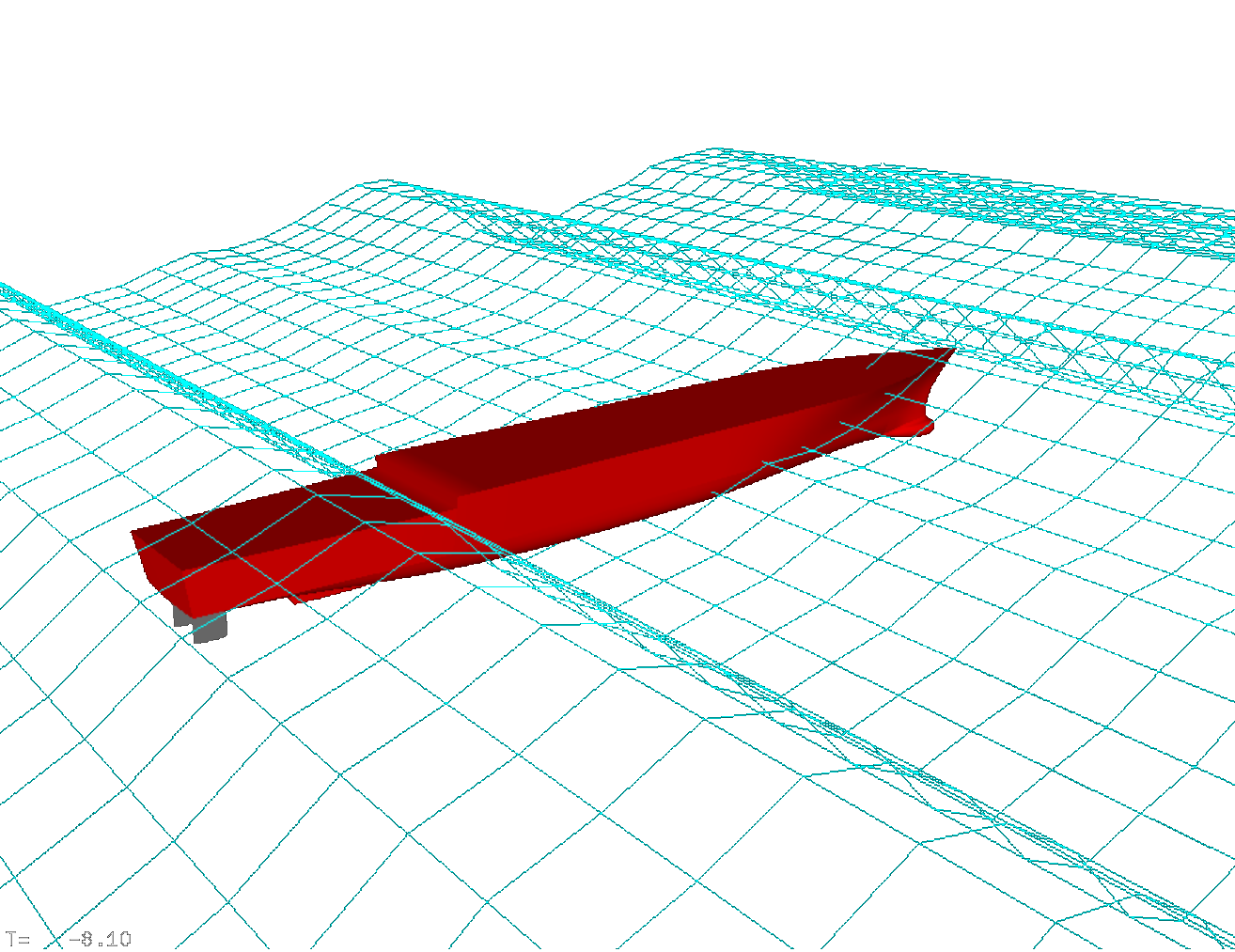}
\includegraphics[width=0.5\linewidth, trim=0 0 0 0, clip] {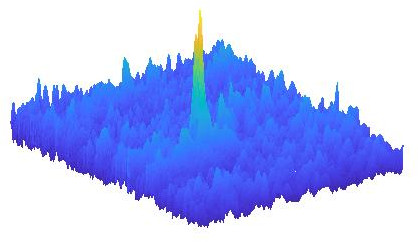}
\caption{\textbf{Left}: Sample visualization of the LAMP simulation.  Here, the vessel passes through extreme waves throwing green water on deck fore and aft.  \textbf{Right}: Sample realization of the MMT model, near an extreme event.}
\label{fig:lamp-mmt}
\end{figure}

For our first benchmark dataset, we use an ocean engineering problem setup described in \cite{guth22} and used previously for benchmarking active learning ideas for extremes in \cite{pickering22a}.  Briefly, the Large Amplitude Motions Program (LAMP) v4.0.9 (May 2019) is used to simulate the structural bending moments of a marine vessel passing through a prescribed wave episode \cite{shin03} \cite{lin07} \cite{lin07b} \cite{lin10}.  In particular, the wave episodes on a finite time interval are defined by the coefficient vector $\vec{\alpha}$ and a reduced order model (ROM) derived from a Karhunen-Lo\`{e}ve (KL) expansion.  Likewise, the Vertical Bending Moment (VBM) time series across that interval is projected onto a low dimensional subspace (derived from principal component analysis).

The ML technique is applied to construct a surrogate model mapping from $\vec{\alpha}$ (the wave episode) to $\vec{q}$ (the resulting VBM).  For practical purposes, we can reconstruct the VBM time series from the $\vec{q}$ using the relationship

\begin{equation}
    M_y(t, \vec{\alpha}) = \sum_{i=1}^{12} q_i(\vec{\alpha}) ~ V_i(t), \ \ \ t \in [0,T]
\end{equation}

\noindent where $V_i(t)$ are the principal components--no more than 12 of these are sufficient to capture the dominant part of the energy \cite{guth22}. In this work we limit ourselves to examining the $\vec{\alpha}$ to $\vec{q}$ step.  Specifically, we restrict our attention to the component $q_1$.

The statistics of the input coefficients are controlled by the ocean sea state, which follows a statistically stationary Gaussian distribution. For this work, we choose a steady sea state corresponding to the JONSWAP spectrum with significant wave height $H_s = 13.2m$ and modal wave period $\omega_m = 10s$ \cite{hasselmann73}.  This is a quite extreme sea state, which  leads to many extreme waves, and result in extreme internal forces, a visualization of which is shown in the left subfigure of figure \ref{fig:lamp-mmt}.  While the choice of spectrum does not \textit{directly} affect the surrogate model, it affects the statistics of the output. In this case the map $q_i( \vec \alpha)$ is non-nonlinear and the resultant statistics highly non-Gaussian.


For each dataset, summarized in table \ref{tab:data-sets}, we first choose a pair $(n, T)$ that describes the KL representation of the wave episodes, where $n$ is the truncation order (dimension of input) and $T$ is the duration of the wave episode.  Then, we perform $n_s$ separate numerical experiments, where the $\vec{\alpha}$ are drawn using Latin Hypercube Sampling (LHS) from an $n$-dimensional hyperbox with interval $[-4.5, 4.5]$.  LHS is a variation of uniform sampling which avoids the density fluctuations common to true (psuedo-) random sampling.

\begin{table}[htpb]
\centering
\begin{tabular}{ c c c c c }
 Input Dimension & Interval Length & Training Set & Validation Set \\
 \hline
 2 & 40 & 25 & 675 \\
 4 & 40 & 462 & 238 \\
 10 & 40 & 2000 & 1000 \\
\end{tabular}
\caption{Selection of LAMP data sets used.}
\label{tab:data-sets}
\end{table}

The hyperbox radius, $z^* = 4.5$, represents $4.5\sigma$ deviations in the choice of the components of $\vec{\alpha}$.  Compared to representative sampling from the steady state, given by
\begin{equation}
    \vec{\alpha} \sim \mathcal{N}(0, \mathbf{I}_n),
    \label{eq:lamp-steady-state}
\end{equation}

\noindent LHS sampling preferentially samples extreme inputs, up to the cutoff $z^* = 4.5$.  For this reason, the LAMP data set includes more extreme wave episodes, and more extreme internal forces, than a data set that would have been sampled from the steady state distribution (equation \ref{eq:lamp-steady-state}).


Finally, each wave episode is equipped with a stochastic prelude, \cite{guth22}.  The stochastic prelude is designed to extrapolate the sea surface elevation outside of the wave episode region, in order to set up the encounter conditions for the vessel.  The stochastic prelude is constructed in a spectrum consistent manner so that the random encounter conditions agree with the steady state conditions while also smoothly matching the wave episode.  Because the stochastic prelude is randomly chosen, it is an important source of aleatoric uncertainty in the relationship between $\vec{\alpha}$ and $\vec{q}$.

We note that the KL expansion provides both a natural ordering of wave episode coefficients, and the spectrum of eigenvalues representing the contribution to the wave episode energy of each dimension.  Thus the $n=2$ data set keeps the two most significant KL modes, while the $n=10$ data set keeps the ten most significant modes.

\subsection{Dispersive wave turbulence data set (MMT data set)}
\label{sec:mmt-data}
The Majda-McLaughlin-Tabak (MMT) model, first introduced in \cite{majda97}, is 1D nonlinear model of deep water wave dispersion.  It has been studied in the context of weak turbulence and intermittent extreme events \cite{cai01, zakharov01, zakharov04, rumpf05, cousins14, guth19, pickering22a}.  A sample visualization of an extreme event in the MMT model is shown in the right subfigure of figure \ref{fig:lamp-mmt}.  The governing equation is given by
\begin{equation}
        iu_t = |\partial_x|^{\alpha_m} u + \lambda |\partial_x|^{\frac{-\beta}{4}} 
        \big( \big| |\partial_x|^{\frac{-\beta}{4}} u \big|^2 |\partial_x|^{\frac{-\beta}{4}} u \big) + iDu,
\end{equation}
\noindent where $u$ is a complex scalar, the pseudodifferential operator $|\partial_x|^{\alpha_m}$ is defined through the Fourier transform as follows:
\begin{equation*}
        \widehat{|\partial_x|^\alpha u(k)} = |k|^\alpha \widehat{u(k)},
\end{equation*}
\noindent exponents $\alpha_m$ and $\beta$ are chosen model parameters, and $D$ is a selective Laplacian.  Our parameter selection is taken from \cite{pickering22a}, and our implementation follows \cite{cousins14}.  Briefly, the spatial domain is periodic on $[0, 1)$ discretized into $512$ points.  The parameters are chosen so that $\lambda = -4$ (focusing case), $\alpha = \frac{1}{2}$ (deep water waves case), and $\beta = 0$.

Our data set is constructed by specifying a (complex) initial condition, according to 
\begin{equation}
    u(x, t=0) = \sum_{j=1}^m \alpha_j \sqrt{\lambda_j} \phi_j(x)
\end{equation}
\noindent where $\vec{\alpha} \in \mathbb{C}^m$ is a vector of complex coefficients with zero mean and unit covariance, the basis vectors $\phi_j(x)$ are the first $m$ eigenvectors of a KL expansion, and the normalization coefficients $\lambda_j$ are the corresponding eigenvalues. This KL expansion is taken from a Gaussian stochastic process over $x \in [0, 1]$, whose autocorrelation is given by the complex periodic kernel 

\begin{equation}
\label{eq:mmt-autocorr}
    k(x, x') = \sigma_u^2 ~ \exp \left(i 2 \sin^2(\pi (x - x')\right) ~ \exp \left(  \frac{-2 \sin^2(\pi (x - x'))}{l_u^2}\right)
\end{equation}

\noindent with $\sigma_u^2 = 1$ and $l_u = 0.35$.  We present brief remarks about this process below in section \ref{sec:compare-data-sets}. We further note that we split the $m$-dimensional complex $\vec{\alpha}$ into a $2m$-dimensional real $\vec{\alpha}$ corresponding to the real and imaginary parts.  We further restrict the $\vec{\alpha}$ to lie within a hyperbox of radius $z^*=6$ (c.f. $z^*=4.5$ above in section \ref{sec:lamp-data}). Finally, we define the output map as
\begin{equation}
    y(\vec{\alpha}) = ||\mbox{Re}\left(u(x, T=50; ~ \vec{\alpha}\right))||_{\infty},
\end{equation}
\noindent that is, the output of the map is the most extreme (real) $u$ in the domain after $T=50$ time units have elapsed.  A table of the data sets considered in this paper is given in table \ref{tab:data-sets-mmt}.

\begin{table}[htpb]
\centering
\begin{tabular}{ c c c c c }
 Input Dimension &  Training Set & Validation Set \\
 \hline
 2 & 25 & 975 \\
 4 & 462 & 99538 \\
 20 & 10000 & 90000 \\
\end{tabular}
\caption{Selection of MMT data sets used.}
\label{tab:data-sets-mmt}
\end{table}

\subsection{Comparison of Data Sets}
\label{sec:compare-data-sets}

\begin{figure}[]
\centering
\includegraphics[width=0.32\linewidth, trim=0 0 0 0, clip] {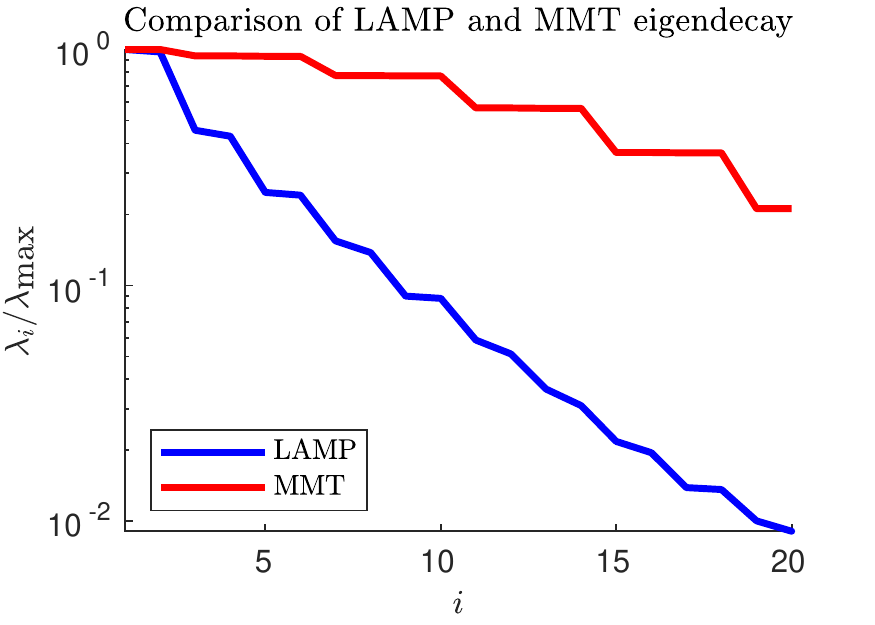}
\includegraphics[width=0.32\linewidth, trim=0 0 0 0, clip] {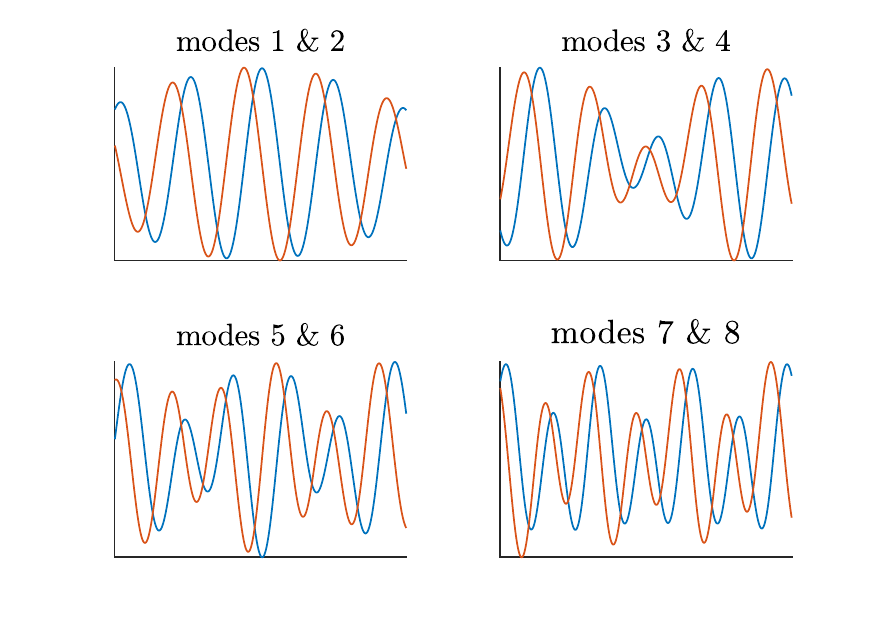}
\includegraphics[width=0.32\linewidth, trim=0 0 0 0, clip] {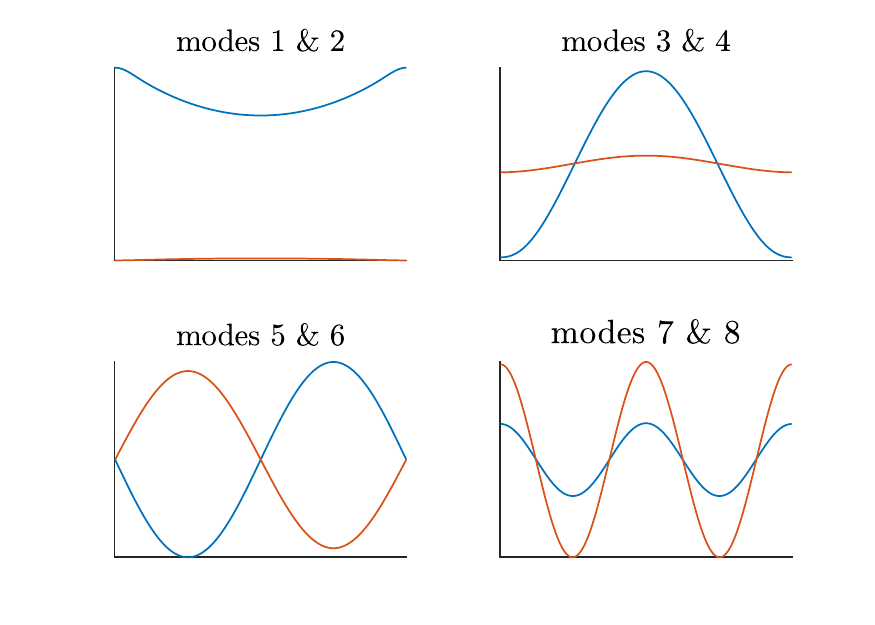}
\caption{\textbf{Left}:  Comparison of the eigenspectral decay for LAMP and MMT reduced order models.  \textbf{Center}: First few LAMP wave episode eigenmodes.  \textbf{Right}: First few MMT initial condition eigenmodes, real parts.  Eigenmodes are presented in pairs to emphasize approximate symmetries.}
\label{fig:eigenspectra}
\end{figure}

\begin{figure}[]
\centering
\includegraphics[width=0.45\linewidth, trim=0 0 0 0, clip] {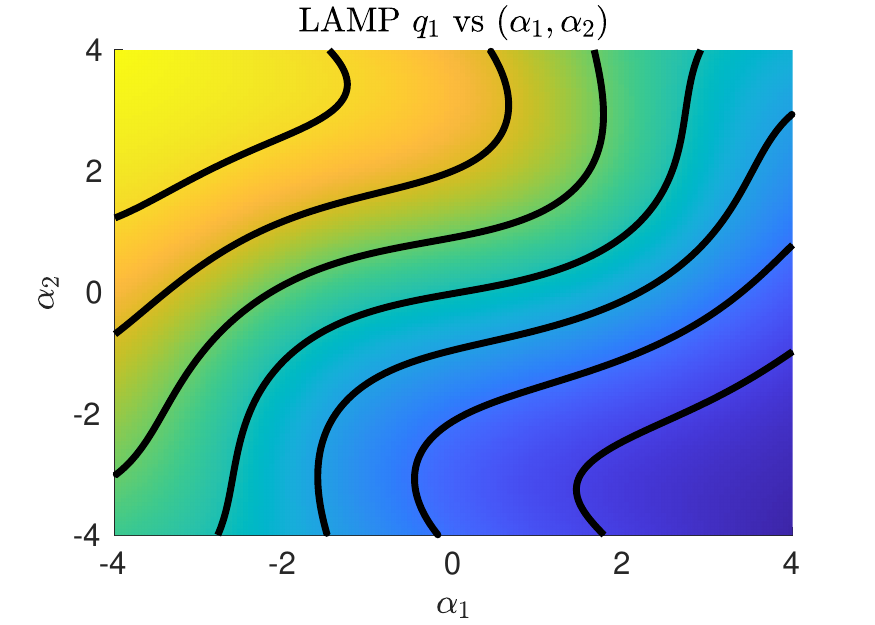}
\includegraphics[width=0.45\linewidth, trim=0 0 0 0, clip] {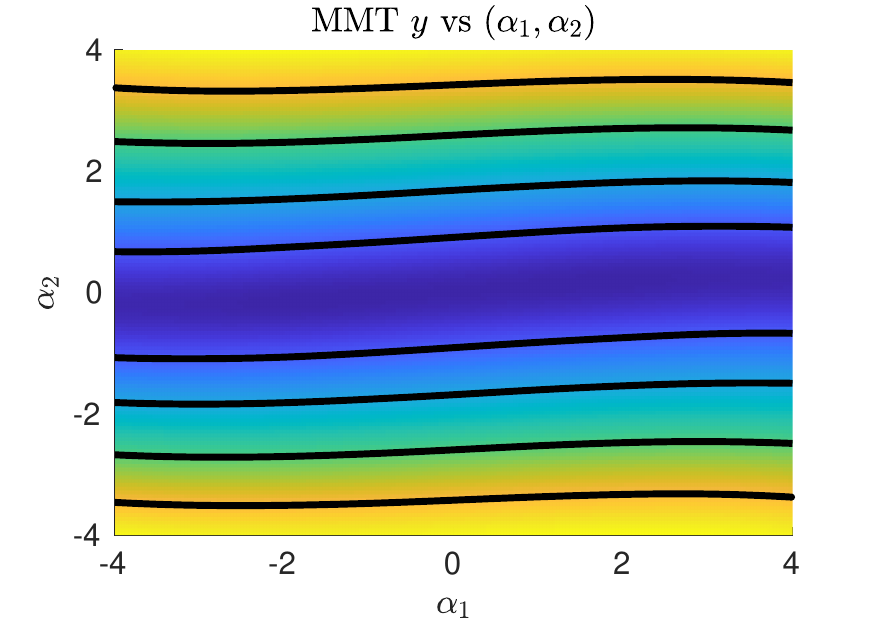}
\caption{\textbf{Left}: Plot of relationship between $(\alpha_1, \alpha_2)$ and $\mathbb{E}\left[q_1\right]$ for the LAMP data set.  \textbf{Right}: Plot of the relationship between $(\alpha_1, \alpha_2)$ and $y$ for the MMT data set.  Both plots are slices, i.e., all other inputs ($\alpha_3, ...$) are zero.}
\label{fig:data-surr-models}
\end{figure}

We briefly note a few features of the two data sets.  In the left subplot of figure \ref{fig:eigenspectra}, we compare the eigenspectrum decay of the reduced order models used in the LAMP dataset and the MMT dataset.  This eigenspectrum describes the energy associated with each mode in the input space.  For the LAMP dataset, this corresponds the KL decomposition of a JONSWAP spectrum with $H_s=13.2m$ and $w_m = 10s$ \cite{hasselmann73, sclavounos12}.  For the MMT dataset, this corresponds to the KL decomposition of the autocorrelation kernel given in equation \ref{eq:mmt-autocorr}. As a preliminary matter, we note that the even/odd symmetry leads to paired eigenvalues for the LAMP input spectrum.  For the MMT spectrum, there is both an even/odd symmetry and an (approximate) real/imaginary symmetry, leading to fourfold repeated eigenvalues.

When we compare the two eigenspectra from figure \ref{fig:eigenspectra}, it is clear that the eigenspectrum decays faster for the LAMP dataset than the MMT dataset.  The faster decay implies that higher order modes are less important in the LAMP data set compared with the MMT data set.  This implies that the even with 10D data, the LAMP dynamics are fundamentally lower dimensional than the MMT dynamics. In the right two subplots of figure \ref{fig:eigenspectra}, we show (the real parts of) the first eight eigenmode shapes for the LAMP and MMT datasets.  The most important feature to note is that the characteristic wavelength of the low order modes for the LAMP problem are shorter than the interval length--approximately four wave periods fit into the interval. This is related to the parameters (peak frequency) of the chosen sea spectrum. For the MMT problem, the characteristic wavelength of the low order modes are longer than the interval. Also, since the autocorrelation function is defined on a periodic domain, the resulted eigenmodes are Fourier modes. 

Next, the two data sets differ in the presences of aleatoric uncertainty, and intrinsic measurement error in the data.  The MMT data is the product of a deterministic 1D PDE, whose only noise source is due to the limits of machine precision.  The LAMP data, however, has stochastic encounter conditions associated with the stochastic prelude technique and a particular wave spectrum \cite{guth22}.  The stochastic encounter conditions manifest as an approximately homoskedastic aleatoric noise term in the in the $q_1$ data.

Besides the energy spectrum of the input side of the data and the aleatoric error, these two problems also differ in the shape of the output side of the data.  In figure \ref{fig:data-surr-models}, we present a slice representation of each data set, where the first two input components ( $\alpha_1, \alpha_2$ ) are varied and the resulting output ($\mathbb{E}\left[q_1\right]$ for LAMP, $y$ for MMT) are plotted.  We see at once that the spatial features in the model are very different.  In particular, the LAMP data has very complicated joint dependence on the different components $\alpha_1$ and $\alpha_2$.  In contrast, the relationship for the MMT data is somewhat more varied, and lacks the `leveling off' seen in the LAMP slice.  This behavior carries through the output statistics--$p_Q(q_1)$ has short tails, while $p_Y(y)$ has a very long right tail.


%
%





\section{Machine Learning Architectures}
\label{sec:ml-arch}

\subsection{Gaussian Process Regression}

\begin{figure}[htpb]
\centering
\begin{tikzpicture}[node distance = 0.5cm, thick,
roundnode/.style={circle, draw=green!60, fill=green!5, very thick, minimum size=12mm},
squarednode/.style={rectangle, draw=red!60, fill=red!5, very thick, minimum size=12mm}
]%

		\node[roundnode, align=center] (x0) {$\vec{\alpha}$};
        \node[squarednode, align=center] (k0) [below=of x0] { kernel \\ matrices };
        \node[roundnode, align=center] (x2) [left=of k0] {$\Theta$, $\sigma_n$};
        \node[squarednode, align=center] (gp) [below=of k0] {$\mathcal{N}(\mu, \sigma)$};
        \node[squarednode, align=center] (nll) [left=of x2] {Negative \\ Log \\ Likelihood};
        \node[roundnode, align=center] (x1) [above=of x2] {$\mathcal{D}$};

        \draw[->] (x0.south) to [out=270,in=90] (k0.north);
        \draw[->] (x1.south) to [out=270,in=90](k0.north);
        \draw[->] (x1.south) to [out=270,in=90](nll.north);
        \draw[->] (x2.east) to (k0.west);
        \draw[->] (k0.south) to [out=270,in=90] (gp.north);
        
        \draw[->] (nll.east) to (x2.west);
        \draw[->] (gp.west) to [out=180,in=270] (nll.south);
        
\end{tikzpicture}%
\caption{Flowchart model of Gaussian Process.}
\label{fig:flowchart-gp}
\end{figure}
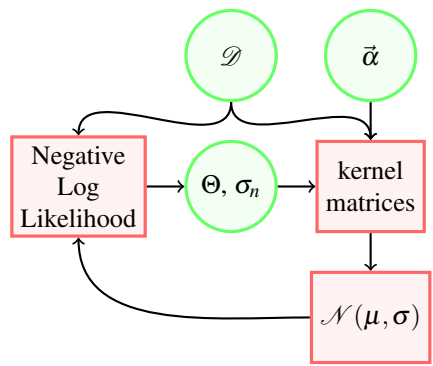

The gold standard for uncertainty quantification in non-parametric regression is Gaussian Processes (GP).  In this method, we use the training data to learn the hyperparameters of a kernel function $k(\vec{\alpha}_1, \vec{\alpha}_2; \theta)$, which describes the statistical relationship between model outputs for two different model inputs.  The choice of kernel function, and especially its smoothness, is an important component of any GP model \cite{genton02}.  For this work, we use the squared exponential kernel with Automatic Relevance Determination (ARD), given by
\begin{equation}
\label{eq:kernel}
    k(\vec{\alpha}_1, \vec{\alpha}_2; \theta) = \theta_{\sigma} \exp\left(-\frac{1}{2}\sum_{i=1}^d\frac{(\alpha_{1,i} - \alpha_{2,i})^2}{\theta_i^2}\right).
\end{equation}
The hyperparameters for GP include both the kernel parameters from equation \ref{eq:kernel} (a variance parameter $\theta_{\sigma}$ and length scales $\theta_i$) as well as the intrinsic noise parameter $\sigma_n$.  The hyperparameter $\sigma_n$ can be understood as a measure of aleatoric measurement error, or alternately as a numerical regularizer as from ridge regression.  These hyperparameters are optimized by minimizing the Negative Log Likelihood (NLL), sometimes called negative marginal log likelihood, given by

\begin{equation}
\label{eq:loss-nll}
    -\log p(\mathbf{y}|\mathbf{A}) = \frac{1}{2}\mathbf{y}^\intercal (K + \sigma_n^2 \mathbf{I} )^{-1}\mathbf{y} + \frac{1}{2} \log |K + \sigma_n^2 \mathbf{I}| + \frac{n}{2} \log 2 \pi
\end{equation}
Once we have optimized the hyperparameters, we calculate the posterior distribution with the relation

\begin{equation}
\label{eq:posterior-dist}
    p(y ~ | ~ \vec{\alpha}) ~ \sim ~ \mathcal{N}\left(\mu(\vec{\alpha}), ~ \sigma(\vec{\alpha})\right)
\end{equation}

\noindent where,

\begin{align}
\label{eq:gp-posterior-noise-1}
        \mu(\vec{\alpha}) & = K_*^\intercal (K + \sigma_n^2 I)^{-1} Y \\
        \sigma^2(\vec{\alpha}) & = K_{**} -  K_*^\intercal (K + \sigma_n^2 I)^{-1} K_*
\label{eq:gp-posterior-noise-2}
\end{align}
\noindent $\mu(\vec{\alpha})$ is the posterior mean, $\sigma^2(\vec{\alpha})$ is the posterior variance, $K$, $K_*$, and $K_{**}$ are kernel matrics, and $Y$ is the observed output data vector (for LAMP, the $q_i$, for MMT, the $y$).  Details of these calculations can be found in standard references, such as Rasmussen and Williams \cite{rasmussen05}.

The advantages of GP for surrogate modeling are twofold.  First, because GP calculates a posterior distribution, it has built in UQ.  Second, the internal structure of the variance calculation provides as simple decomposition of the uncertainty into an aleatoric term ($\sigma_n$), and an epistemic term (the remainder).

On the other hand, GP has two major disadvantages.  First, the matrix math of exact GP calculations involves inverting a matrix whose size scales like the number of training samples.  This $O(n_s^3)$ computational cost scaling prohibits the use of GP for large data sets ($n_s \gtrsim 1000$), at least without significant approximations.  Second, exploration of the the kernel function space is difficult for high dimensional spaces.  As a result, GP models struggle with high dimensional input spaces.  Because GP is restricted away from high dimensional input, we do not consider GP in conjuction with the functional input described below in section \ref{sec:neural-op}.

\subsection{Traditional Neural Networks}

\begin{figure}
    \centering
\begin{tikzpicture}[node distance = 0.5cm, thick,
roundnode/.style={circle, draw=green!60, fill=green!5, very thick, minimum size=12mm},
squarednode/.style={rectangle, draw=red!60, fill=red!5, very thick, minimum size=12mm}
]%

		\node[roundnode, align=center] (x0) {$\vec{\alpha}$};
		\node[squarednode, align=center] (k0) [below=of x0] {NN};
		\node[squarednode, align=center] (x1) [left=of k0] {trained \\ weights};
        \node[squarednode, align=center] (mse) [left=of x1] {mean \\ squared \\ error};
		\node[roundnode, align=center] (d0) [above=of mse] {$\mathcal{D}$};
        \node[roundnode, align=center] (mu) [below=of k0] {$y$};

        \draw[->] (x0.south) -- (k0.north);
        \draw[->] (x1.east) -- (k0.west);
        \draw[->] (k0.south) -- (mu.north);
        \draw[->] (d0.south) -- (mse.north);
        \draw[->] (mse.east) -- (x1.west);
        \draw[->] (mu.west) to [out=180,in=270]  (mse.south);
\end{tikzpicture}%

    \caption{Flowchart model of a traditional feed-forward Neural Network.}
\label{fig:flowchart-nn}
\end{figure}
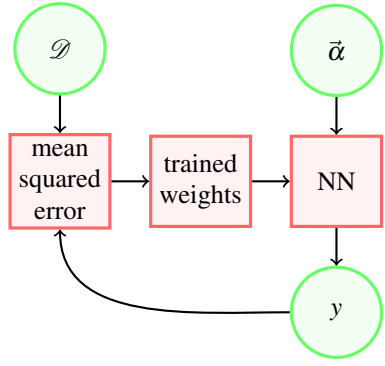

In this section we briefly recapitulate some of the features of a traditional feed-forward neural network.  During the inference step, computation passes from the input layer (here, $\vec{\alpha}$) through a number of a hidden layers.  At each hidden layer, the values from the previous layer are multiplied by a weight matrix, and then fed into a nonlinear activation function.  Common activation functions include the sigmoid and the hyperbolic tangent.  In this paper, we use the rectified linear unit (ReLU).  At the final (output) layer, the values from the last hidden layer are combined in a weighted sum, and a single output is reported.

In the simplest case, the input, weights, and activation functions are all deterministic, so the output from a feed forward neural network is also deterministic.  The values of the weights parameters, however, are determined in a separate training step.  During training, the model iterates through a training set, and compares model predicted output to the training data output according to some loss function, such as mean squared error.  Model errors are used to adjust the network weights through the backpropagation algorithm.  The literature on neural network training is vast, and we direct the interested reader to a reference such as \cite{shalev-shwartz14} or \cite{murphy22}.  We note that, unlike the inference (feed-forward) step, network training is generally a non-deterministic stochastic gradient descent (SGD) process.

In figure \ref{fig:flowchart-nn} we display a schematic flowchart of the relationship between the network input ($\vec{\alpha}$), output ($y$), and weights, which in turn depend on the training data ($\mathcal{D}$).  The important architecture hyperparameters for a feed-forward neural network are the number of hidden layers, the size of each hidden layer, and the choice of activation function.  Additionally, important training hyperparameters include the training time (epochs), the training rate (automatically adjusted by the Adams gradient descent), and the batch size (division of data during training).  Additionally, regularization techniques, such as dropout or $L_2$ Tikhonov regularization, may be used to combat overfitting. We emphasize that, unlike the GP case, the simple neural network produces a single (scalar) output prediction instead of a distribution.  This form of prediction is insufficient for UQ, so we next turn to a series of techniques to modify neural networks in order to allow for UQ.

\subsection{Gaussian Neural Networks}

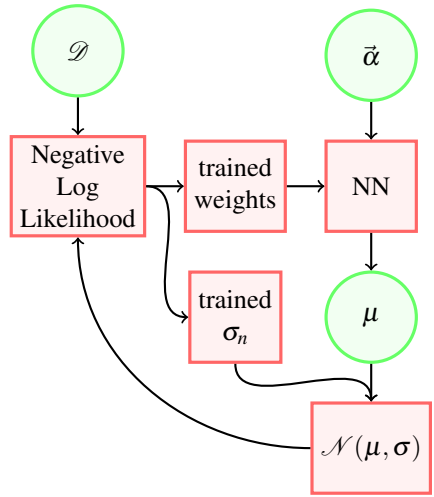
\begin{figure}
    \centering
    \begin{tikzpicture}[node distance = 0.5cm, thick,
roundnode/.style={circle, draw=green!60, fill=green!5, very thick, minimum size=12mm},
squarednode/.style={rectangle, draw=red!60, fill=red!5, very thick, minimum size=12mm}
]%

		\node[roundnode, align=center] (x0) {$\vec{\alpha}$};
		\node[squarednode, align=center] (k0) [below=of x0] {NN};
		\node[squarednode, align=center] (x1) [left=of k0] {trained \\ weights};
        \node[roundnode, align=center] (mu) [below=of k0] {$\mu$};
        \node[squarednode, align=center] (sigma) [below=of x1] {trained \\ $\sigma_n$};
        \node[squarednode, align=center] (gp) [below=of mu] {$\mathcal{N}(\mu, \sigma)$};
        
        \node[squarednode, align=center] (nll) [left=of x1] {Negative \\ Log \\ Likelihood};
		\node[roundnode, align=center] (d0) [above=of nll] {$\mathcal{D}$};

        \draw[->] (x0.south) -- (k0.north);
        \draw[->] (x1.east) -- (k0.west);
        \draw[->] (k0.south) -- (mu.north);
        \draw[->] (mu.south) -- (gp.north);
        \draw[->] (sigma.south) to [out=270,in=90] (gp.north);
        \draw[->] (d0.south) -- (nll.north);
        
        
        \draw[->] (gp.west)  to [out=180,in=270] (nll.south);
        \draw[->] (nll.east) to [out=0,in=180] (sigma.west);
        \draw[->] (nll.east) to [out=0,in=180] (x1.west);
\end{tikzpicture}%
    \caption{Flowchart model of Gaussian Neural Network.}
\label{fig:flowchart-gnn}
\end{figure}


The first method of introducing uncertainty quantification to neural networks takes direct inspiration from GP, while attempting to avoid the iron $O(n_s^3)$ cost of matrix inversion.  Wilson et al (2015) \cite{wilson15} developed KISS-GP with an approximate kernel matrix for fast inversion and determinant calculations.  Calandra et al (2016) \cite{calandra16} developed Manifold GP as a technique to learn an encoder and latent space representation instead of kernel function hyperparameters.  Burt et al. (2020) \cite{burt20} investigated using variational inference instead of an exact kernel to speed matrix inversions.

The common point of these approaches is Inducing Point GP or Deep Kernel Learning, also investigated by Ober and Rasmussen (2021) \cite{ober21a} and  Amersfoort et al. (2021) \cite{amersfoort20, amersfoort21}.   The common idea is to take a final Gaussian Process `layer,' but replace the kernel matrix calculations with a different approximate learning scheme.  By retaining the posterior distribution, we are able to keep the built-in UQ from GP based techniques.


In figure \ref{fig:flowchart-gnn}, we show a schematic representation of our implementation, which we call Gaussian Neural Network (G-NN).  Starting from equation \ref{eq:posterior-dist} for the posterior distribution, we train a neural network to learn the function $\mu(\vec{\alpha})$.  Simultaneously, we train a parameter for the homoskedastic variance term $\sigma_n$.  Because the G-NN outputs a distribution instead of a single point, we use the same (NLL) loss function for training as GP, given in equation \ref{eq:loss-nll}. The G-NN model fits a homoskedastic noise parameter $\sigma_n$ to the data set, in a close match to how GP estimates the intrinsic noise parameter $\sigma_n$.  However, unlike, GP, G-NN does not estimate epistemic portion of the posterior variance, which could vary with $\vec{\alpha}$.  As a result, while we can interpret the G-NN $\sigma_n$ as aleatoric uncertainty, G-NN does not predict the epistemic uncertainty.

We note that, like the $\sigma_n$ term in GP, it is possible to learn a more complicated function form for the intrinsic noise term in G-NN.  However, overfitting $\sigma_n$ is a distinct danger.  Additionally, for the ocean engineering dataset we examine, a homoskedastic aleatoric uncertainty model is a reasonable approximation.  For this reason, we do not evaluate a heteroskedastic G-NN here. We note also that, in our experience, training a G-NN on a noiseless dataset introduces numerical convergence complications and is not recommended.  For those cases, such as the MMT dataset here, we recommend using a traditional deterministic output layer.

\subsection{Prediction Ensembles}


One of the most used frameworks for introducing UQ into NN models is Variational Inference (VI) \cite{blei17}.  Under this framework, the latent variables of the model (network weights) have some conditional distribution given the observed training data.  Forward inference from these exact conditional distributions would produce a predicted output distribution, whose spread would include the model epistemic uncertainty.

As computing the exact conditional distributions is computationally intractable, the literature has proposed methods to approximately sample model realizations to produce the Bayesian Model Average \cite{psaros23}.  In this paradigm, we construct an ensemble of neural networks, each of which produces a single prediction.  We then approximate the true posterior distribution with the ensemble of predictions, using the relation (written for convenience for a scalar output)
\begin{equation}
    p(y ~ | ~ \vec{\alpha}, \mathcal{D}) \approx \frac{1}{n_e} \sum_{i=1}^{n_e} p(y ~ | ~ \vec{\alpha}, \hat{\theta}_i),
\end{equation}

\noindent where $\mathcal{D}$ is the observed training data, $n_e$ is the ensemble size, and $\hat{\theta}_i$ is the parametrization (weights) of the $i^{\mbox{th}}$ network. In the simple case where each network produced a single prediction $\{y_i\}$, we compute the ensemble variance, given by 

\begin{equation}
    \sigma^2_{\mbox{ens}} = \frac{1}{n_e -1}\sum_{j=1}^{n_e} (\overline{y} - y_j)^2,
\end{equation}

\noindent and use it as a proxy for the epistemic variance: $\sigma^2_{\mbox{ens}} \approx \sigma^2_{\epsilon}$.  

In the following sections, we present three techniques to produce an ensemble of predictions:  Ensemble Neural Networks (ENN), Bayesian Neural Networks (BNN), and Dropout Neural Networks (D-NN).  For each technique, we explain how to generate an ensemble of predictions, and how to identify the parameter $n_e$ which controls the ensemble size. Finally, we emphasize that the ensemble technique is fully compatible with the G-NN technique.  That is, we can use both an ensemble technique and the Gaussian process layer to estimate aleatoric and epistemic uncertainties simultaneously.  When we combine these techniques, we produce an ensemble of Gaussian distributions $\mathcal{N}(\mu_i, \sigma_{n,i})$.  From this ensemble, we estimate 
\begin{equation}
    \sigma^2_n = \frac{1}{n_e}\sum_{i=1}^{n_e} \sigma^2_{n,i}, \qquad \sigma^2_{\epsilon} = \mbox{Var}(\{\mu_i\}),
\end{equation}
\noindent in a decomposition based on the law of total variance.

\subsubsection{Ensemble Neural Networks}

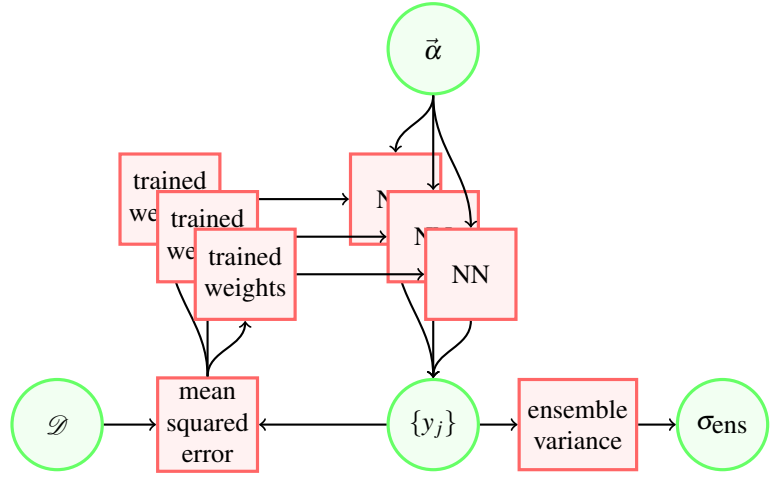
\begin{figure}[htpb]
    \centering
    \begin{tikzpicture}[node distance = 0.5cm, thick,
roundnode/.style={circle, draw=green!60, fill=green!5, very thick, minimum size=12mm},
squarednode/.style={rectangle, draw=red!60, fill=red!5, very thick, minimum size=12mm}
]%

		\node[roundnode, align=center] (x0) at (4.5, 6) {$\vec{\alpha}$};
        \node[roundnode, align=center] (d0) at (-0.5, 1) {$\mathcal{D}$};
        \node[roundnode, align=center] (mu) at (4.5, 1) {$\{y_j\}$};
        \node[squarednode, align=center] (ens) [right=of mu] {ensemble \\ variance};
        \node[roundnode, align=center] (sig) [right=of ens] {$\sigma_{\mbox{ens}}$};
        
        \node[squarednode, align=center] (mse) at (1.5, 1) {mean \\ squared \\ error};

		\node[squarednode, align=center] (nn1) at (4, 4) {NN};
        \node[squarednode, align=center] (w1) at (1, 4) {trained \\ weights};
        
        \draw[->] (x0.south) to [out=270,in=90] (nn1.north);
        \draw[->] (nn1.south) to [out=270,in=90] (mu.north);
        \draw[->] (mse.north) to [out=90,in=270] (w1.south);
        \draw[->] (w1.east) -- (nn1.west);
        
		\node[squarednode, align=center] (nn2) at (4.5, 3.5) {NN};
        \node[squarednode, align=center] (w2) at (1.5, 3.5) {trained \\ weights};

        \draw[->] (x0.south) to [out=270,in=90] (nn2.north);
        \draw[->] (nn2.south) to [out=270,in=90] (mu.north);
        \draw[->] (mse.north) to [out=90,in=270] (w2.south);
        \draw[->] (w2.east) -- (nn2.west);
  
		\node[squarednode, align=center] (nn3) at (5, 3) {NN};
		\node[squarednode, align=center] (w3) at (2, 3) {trained \\ weights};
  
        \draw[->] (x0.south) to [out=270,in=90] (nn3.north);
        \draw[->] (nn3.south) to [out=270,in=90] (mu.north);
        \draw[->] (mse.north) to [out=90,in=270] (w3.south);
        \draw[->] (w3.east) -- (nn3.west);

        \draw[->] (mu.east) -- (ens.west);
        \draw[->] (ens.east) -- (sig.west);
        
        \draw[->] (mu.west) -- (mse.east);
        \draw[->] (d0.east) -- (mse.west);

\end{tikzpicture}%
    \caption{Flowchart model of Ensemble Neural Network (ENN).}
\label{fig:flowchart-enn}
\end{figure}

The first technique for creating a prediction ensemble is the Ensemble Neural Network (ENN), the implementation of which we borrow from an operator network investigation by Pickering et al. (2022) \cite{pickering22a}.  The underlying idea of ENN is to train multiple neural networks with different (random) initial weights on the same training data.  Then, during inference, each network is sampled separately, producing an ensemble of predictions.  We simply use the ensemble mean and ensemble variance of these predictions as parameters of a Gaussian posterior distribution $p(y_i ~ | ~ \vec{\alpha})$.  This architecture is shown is schematically in figure \ref{fig:flowchart-enn}. 

Intuitively, this approach has promise.  Near training data, the loss function will enforce that each network in the ensemble agree with one another, because they must agree with the training data.  In regions far from the training data, however, there is no such pressure, and the ensemble predictions are free to diverge. Unlike the other approaches we describe, the ENN architecture has no built-in regularization term.  For our results, we include an $L_2$ weight regularizer. 

\begin{figure}[htpb]
\centering
\includegraphics[width=0.45\linewidth, trim=0 0 0 0, clip] {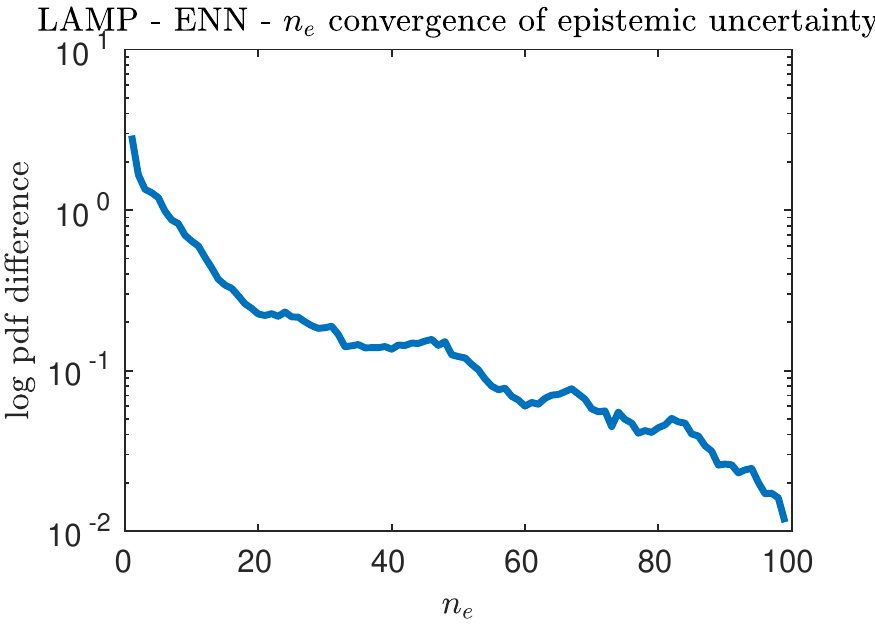}
\includegraphics[width=0.45\linewidth, trim=0 0 0 0, clip] {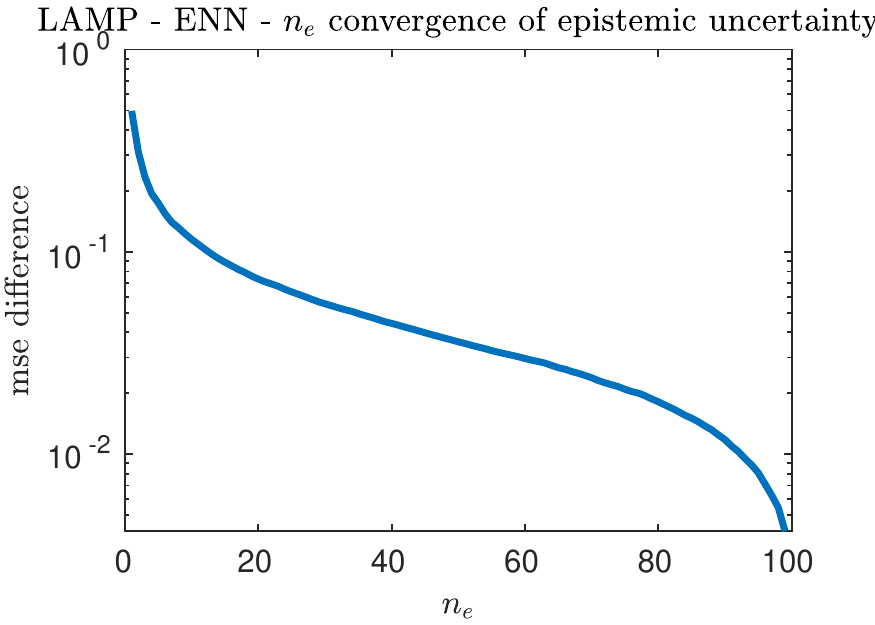}
\caption{Convergence curves of $\sigma_\epsilon$ as functions of ensemble size, $n_e$, for ENN surrogate, $N=10, n_s=2000$. \textbf{Left}: log pdf difference: $\int{| \log{p_S(\sigma_\epsilon^{n_e})} - \log{p_S(\sigma_\epsilon^{100})} |\text{d}\sigma_\epsilon}$.  \textbf{Right}: Mean of squared differences: $\sum{(\sigma_\epsilon^{n_e} - \sigma_\epsilon^{100}})^2/n_s$.}
\label{fig:enn-ne}
\end{figure}

\noindent \textbf{Ensemble Size}. Because each neural network must be trained separately, training time is proportional to the ensemble size, $n_e$.  Pickering et al. \cite{pickering22a} has suggested that just two networks ($n_e = 2$) was sufficient to estimate relative uncertainties for active sampling.  We found that a larger ensemble -- up to $n_e = 8$ -- gave better estimates for more robust uncertainty quantification. In particular, the hyperparameter $n_e$ trades training time and memory costs against statistical power for inferring the true ensemble-distribution variance from the ensemble-sample variance.  As $n_e$ increases, the accuracy of the ensemble variance improves.  However, training time, and memory cost, is directly proportional to $n_e$.  This cost is particularly acute in active learning applications, where many surrogates must be trained iteratively.  Conversely, the computational cost is somewhat ameliorated by the ease of parallelization--each member of the ensemble may be trained independently.

In figure \ref{fig:enn-ne}, we show how the distribution of the model uncertainty, $p_s(\sigma_\epsilon)$, i.e. the histrogram  all values of the estimated epistemic uncertainty over the input parameter space, vary with the ensemble size, $n_e$. We use this quantity as a collective measure for the overall shape of $\sigma_\epsilon$ for high dimensional problems.  For the 10D LAMP dataset, we run a parameter convergence study for the steady state statistics (left subplot, log pdf difference) and the pointwise epistemic errors (right subplot, mse).  We note that, after an initial period of rapid convergence, the improvement for increasing $n_e$ rapidly levels off (the fast drop near $n_e=100$ is an artifact of using $n_e=100$ as the convergence target). We conclude that, for ENN, there is a significant improvement from $n_e=2$ to $n_e=3$, and noticeable improvements perhaps through $n_e = 10$, though the benefit is paid for by a linear increase in training (and inference) cost.  For all ENN results below, we fix $n_e=8$.




\subsubsection{Bayesian Neural Networks}

\begin{figure}[htpb]
\centering
\begin{tikzpicture}[node distance = 0.5cm, thick,
roundnode/.style={circle, draw=green!60, fill=green!5, very thick, minimum size=12mm},
squarednode/.style={rectangle, draw=red!60, fill=red!5, very thick, minimum size=8mm}
]%

		\node[roundnode, align=center] (x0) {$\vec{\alpha}$};
		\node[squarednode, align=center] (n0) [below=of x0] {Fully Connected};
		\node[squarednode, align=center] (n1) [below=of n0] {$250 ~ \times$ ReLU};
		\node[squarednode, align=center] (n2) [below=of n1] {Fully Connected};
		\node[squarednode, align=center] (n3) [below=of n2] {$250 ~ \times$ ReLU};
		\node[squarednode, align=center] (n4) [below=of n3] {Fully Connected};
        
        \node[roundnode, align=center] (d1) [left=of n0] {$\mathcal{N}(\textbf{w}_1, \Sigma_1)$};
        \node[roundnode, align=center] (d2) [left=of n2] {$\mathcal{N}(\textbf{w}_2, \Sigma_2)$};
        \node[roundnode, align=center] (d3) [left=of n4] {$\mathcal{N}(\textbf{w}_3, \Sigma_3)$};
	
        \node[roundnode, align=center] (mu) at (0, -9) {$\{y_i\}$};
        
        \node[squarednode, align=center] (posterior_opt) [left=of d2] {parameters \\ trained via \\ Variational \\ Inference};
        \node[squarednode, align=center] (elbo) at (-5.2, -9) {-ELbo loss: \\ $\mathcal{L} = -\mathcal{F}$};
        \node[roundnode, align=center] (dataset) [left=of elbo] {$\mathcal{D}$};

        \node[squarednode, align=center] (ens) [right=of mu] {ensemble \\ variance};
        \node[roundnode, align=center] (sig) [right=of ens] {$\sigma_{\mbox{ens}}$};

        \draw[->] (x0.south) -- (n0.north);
		\draw[->] (n0.south) -- (n1.north);
		\draw[->] (n1.south) -- (n2.north);
		\draw[->] (n2.south) -- (n3.north);
		\draw[->] (n3.south) -- (n4.north);
        
        \draw[->] (n4.south) -- (mu.north);
        
        \draw[->] (d1.east) -- (n0.west);
        \draw[->] (d2.east) -- (n2.west);
        \draw[->] (d3.east) -- (n4.west);
        
        \draw[->] (mu.east) -- (ens.west);
        \draw[->] (ens.east) -- (sig.west);
        
        
        \draw[->] (dataset.east) -- (elbo.west);
        
        \draw[->] (elbo.north) -- (posterior_opt.south);
        
        \draw[->] (mu.west) -- (elbo.east);
        
        \draw[->] (posterior_opt.east) to [out=0,in=180] (d1.west);
        \draw[->] (posterior_opt.east) -- (d2.west);
        \draw[->] (posterior_opt.east) to [out=0,in=180] (d3.west);
\end{tikzpicture}%
\caption{Flowchart model of Bayesian Neural Network (BNN) with 2 hidden layers.}
\label{fig:flowchart-bnn}
\end{figure}
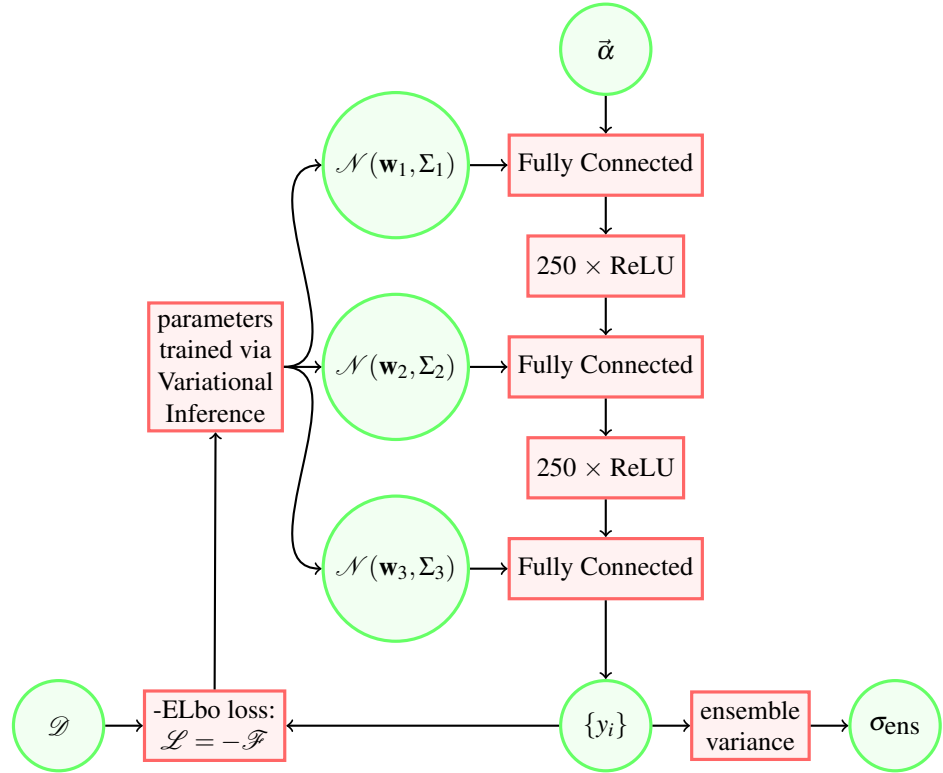

Plain feedforward neural networks are prone to overfitting. When applied to supervised learning problems these networks are also often incapable of correctly assessing the uncertainty in the training data and so make overly confident decisions about the correct class, prediction or action. These issues are addressed by using variational Bayesian learning to introduce uncertainty in the weights of the network. In this approach, all the weights in the neural networks are represented by probability distributions over possible values, rather than having a single fixed value as in traditional neural networks \cite{jospin22, goan20}.  In figure \ref{fig:flowchart-bnn}, we show the relationship of the Bayesian network probability distributions to the traditional neural network weights.

Learnt representations and computations must therefore be robust under perturbation of the weights, but the amount of perturbation each weight exhibits is also learnt in a way that coherently explains variability in the training data. Thus instead of training a single network, essentially an ensemble of networks are trained, where each network has its weights drawn from a shared, learnt probability distribution. Unlike other ensemble methods, this approach typically only doubles the number of parameters yet trains an infinite ensemble using unbiased Monte Carlo estimates of the gradients. In general, exact Bayesian inference on the weights of a neural network is intractable as the number of parameters is very large and the functional form of a neural network does not lend itself to exact integration. Instead we take a variational approximation to exact Bayesian updates. 

\noindent \textbf{Training a Bayesian Neural Network}.
Bayesian inference for neural networks calculates the posterior distribution of the weights given the training data, $P(\textbf{w}|\mathcal D)$. This distribution answers predictive queries about unseen data by taking expectations: the predictive distribution of an unknown label $\hat y$ of a test data item $\hat{\textbf{x}}$, is given by $P(\hat y|\hat{\textbf{x}}) = \mathbb{E}_{P(\textbf{w}|\mathcal D)} [P(\hat y|\hat{\textbf{x}}, \textbf{w})]$. Each possible configuration of the weights, weighted according to the posterior distribution, makes a prediction about the unknown label given the test data item $\hat{\textbf{x}}$. Thus taking an expectation under the posterior distribution on weights is equivalent to using an ensemble of an uncountably infinite number of neural networks. Unfortunately, this is intractable for neural networks of any practical size. To work around this problem, we employ variational learning, i.e., variational learning finds the parameters $\theta$ of a distribution on the weights $q(\textbf{w}|\theta)$ that minimizes the Kullback-Leibler (KL) Weight Uncertainty in Neural Networks divergence with the true Bayesian posterior on the weights. The resulting cost function is known as the variational free energy or the expected lower bound (ELBo) and is expressed as:
\begin{equation} \label{elbo}
    \mathcal{F}(\mathcal{D}, \theta) = \text{\textbf{KL}}[q(\textbf{w}|\theta) \| P(\textbf{w})] - \mathbb{E}_{q(\textbf{w}|\theta)}[\log P(\mathcal{D}|\textbf{w})]
\end{equation}
The cost function of eq. \ref{elbo} is a sum of a data-dependent part, which is referred to as the likelihood cost, and a prior-dependent part, which is referred to as the complexity cost. The cost function embodies a trade-off between satisfying the complexity of the data $\mathcal{D}$ and satisfying the simplicity prior $P(\textbf{w})$. Eq. \ref{elbo} is also readily given an information theoretic interpretation as a minimum description length cost. Exactly minimizing this cost naively is computationally prohibitive. Instead gradient descent and various approximations are used. Due to complexity of the approach and the stochastic nature of the method, the modeling and the optimization is done through the so called "Bayes By Backprop" method. 
For more details on the implementation please see the original article by \cite{blundell2015weight}. After the training is done, for every forward-pass (prediction), weights of the BNN are randomly drawn from $y(\textbf{w})$. Assuming that $y(\textbf{w})$ is a normal distribution, the weights at every forward-pass are:
\begin{equation}
    w_i = w_0^i + \varepsilon \sigma_i, \hspace{10pt} \varepsilon \sim \mathcal{N}(0, 1).
\end{equation}

Consequently, if an input is passed through the BNN several times, the outputs will be slightly different each time and form their own distribution. The mean of these distributions are considered the outputs (predictions) and their standard deviation is the epistemic uncertainty. For the results presented here, we pass each input through the BNN 100 times (ensemble size of 100).  

\subsubsection{Dropout Neural Networks}

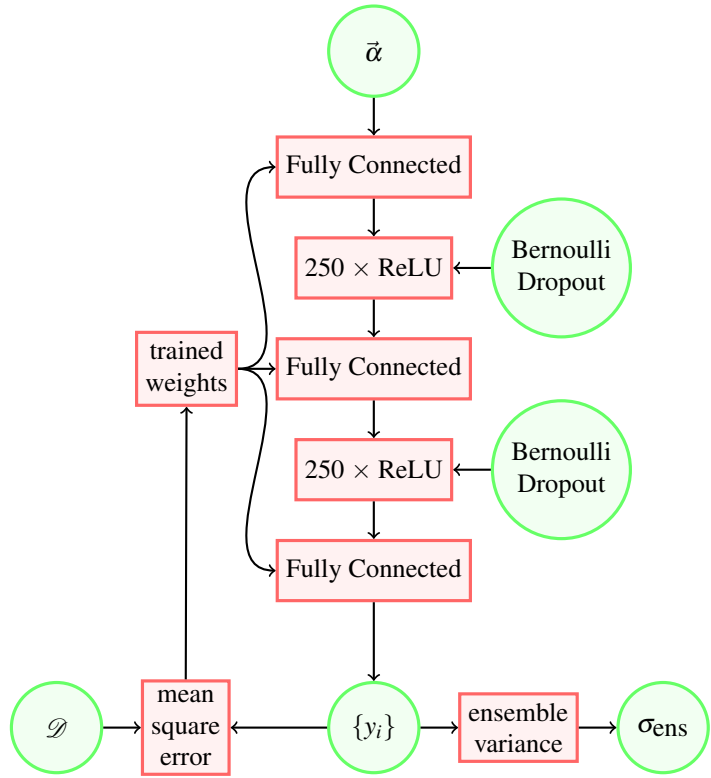
\begin{figure}[htpb]
\centering
\begin{tikzpicture}[node distance = 0.5cm, thick,
roundnode/.style={circle, draw=green!60, fill=green!5, very thick, minimum size=12mm},
squarednode/.style={rectangle, draw=red!60, fill=red!5, very thick, minimum size=8mm}
]%

		\node[roundnode, align=center] (x0) {$\vec{\alpha}$};
		\node[squarednode, align=center] (n0) [below=of x0] {Fully Connected};
		\node[squarednode, align=center] (n1) [below=of n0] {$250 ~ \times$ ReLU};
		\node[squarednode, align=center] (n2) [below=of n1] {Fully Connected};
		\node[squarednode, align=center] (w1) [left=of n2] {trained \\ weights};
		\node[squarednode, align=center] (n3) [below=of n2] {$250 ~ \times$ ReLU};
		\node[squarednode, align=center] (n4) [below=of n3] {Fully Connected};

        \node[roundnode, align=center] (mu) at (0, -9) {$\{y_i\}$};
        
        \node[roundnode, align=center] (d1) [right=of n1] {Bernoulli \\ Dropout};
        \node[roundnode, align=center] (d2) [right=of n3] {Bernoulli \\ Dropout};
        
        \node[squarednode, align=center] (mse) at (-2.5, -9) {mean \\ square \\ error};
        \node[roundnode, align=center] (dataset) [left=of mse] {$\mathcal{D}$};
        
        \node[squarednode, align=center] (ens) [right=of mu] {ensemble \\ variance};
        \node[roundnode, align=center] (sig) [right=of ens] {$\sigma_{\mbox{ens}}$};

        \draw[->] (x0.south) -- (n0.north);
		\draw[->] (n0.south) -- (n1.north);
		\draw[->] (n1.south) -- (n2.north);
		\draw[->] (n2.south) -- (n3.north);
		\draw[->] (n3.south) -- (n4.north);
        
        \draw[->] (n4.south) -- (mu.north);
        
        \draw[->] (d1.west) -- (n1.east);
        \draw[->] (d2.west) -- (n3.east);
        
        \draw[->] (mu.east) -- (ens.west);
        \draw[->] (ens.east) -- (sig.west);
        
        
        
        \draw[->] (dataset.east) -- (mse.west);
        \draw[->] (mu.west) -- (mse.east);
        \draw[->] (mse.north) -- (w1.south);
        
        \draw[->] (w1.east) to [out=0,in=180] (n0.west);
        \draw[->] (w1.east) -- (n2.west);
        \draw[->] (w1.east) to [out=0,in=180] (n4.west);

\end{tikzpicture}%
\caption{Flowchart model of Dropout Neural Network (D-NN) with 2 hidden layers.}
\label{fig:flowchart-dnn}
\end{figure}


\begin{figure}[htpb]
\centering
\includegraphics[width=0.45\linewidth, trim=0 0 0 0, clip] {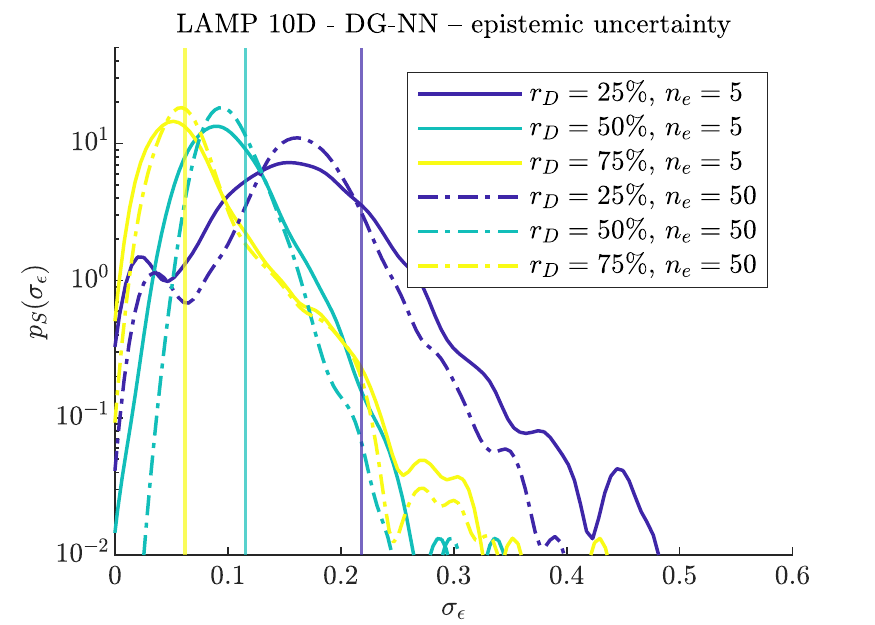}
\includegraphics[width=0.45\linewidth, trim=0 0 0 0, clip] {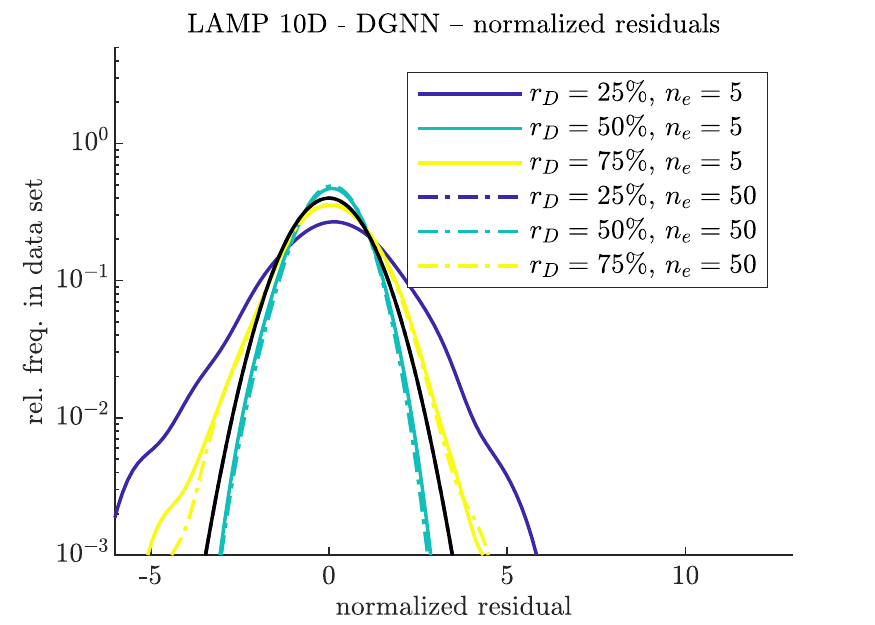}
\caption{Comparison of Gaussian D-NN (DG-NN) surrogate models for different dropout parameters, $r_D$, and ensemble size, $n_e$.  $N=10, n_s=2000$. Additionally, the configuration of the neural network is as described before.  \textbf{Left}:  Distribution of epistemic uncertainties and aleatoric uncertainties (vertical lines).  \textbf{Right}: Distribution of normalized residuals compared against a normal distribution (black solid curve).}
\label{fig:dnn-nd}
\end{figure}

Bayesian probability theory offers mathematically grounded tools to reason about model uncertainty, but these usually come with a prohibitive computational cost. It is perhaps surprising then that it is possible to cast recent deep learning tools as Bayesian models – without changing either the models or the optimization. It can be shown that the use of dropout (and its variants) in NNs can be interpreted as a Bayesian approximation of a well known probabilistic model: the Gaussian process (GP) \cite{damianou13}. Dropout is used in many models in deep learning as a way to avoid over-fitting and this interpretation suggests that dropout approximately integrates over the model weights \cite{srivastava14}. In other words, it is shown that a neural network with arbitrary depth and non-linearities, with dropout applied before every layer, is mathematically equivalent to an approximation
to the probabilistic deep Gaussian process (marginalised over its covariance function parameters). It can also be shown that the dropout objective, in effect, minimizes the Kullback–Leibler divergence between an approximate distribution and the posterior of a deep Gaussian process (marginalised over its finite rank covariance function parameters) \cite{gal2016dropout}. 

The approach we take in this article is that, before each layer of our 8$\times$[250] neural network model (except the first hidden layer), we add a dropout layer with dropout rate of $r_D$. The dropout layer sets neurons (input to the next layer) to zero randomly with probability of $r_D$ at every \textit{training step}, while scaling up the rest (non-zero neuron or inputs) by a factor of $1/(1-r_D)$ so that the sum over all neurons (inputs) remains unchanged. Typically, after the training is done, the dropout layers are switched off which makes the neural network to have deterministic output, however, here, we keep them on during prediction, as well. This means that during \textit{every forward-pass} (prediction) some of the neurons in each hidden layer are randomly \textit{switched off}, i.e., at every forward-pass the prediction is made by a slightly changed NN. This means that the D-NN predicts a different output associated with each input every time it is passed through the D-NN. This leads to D-NN having a probabilistic output.   

We show a schematic representation of the D-NN architecture in figure \ref{fig:flowchart-dnn}, and in particular emphasize the differences between the D-NN and BNN architectures:  the BNN probability distribution controls the \textit{weights}, whereas the Dropout probabilities control the nodes in each hidden layer.

\begin{figure}[htpb]
\centering
\includegraphics[width=0.45\linewidth, trim=0 0 0 0, clip] {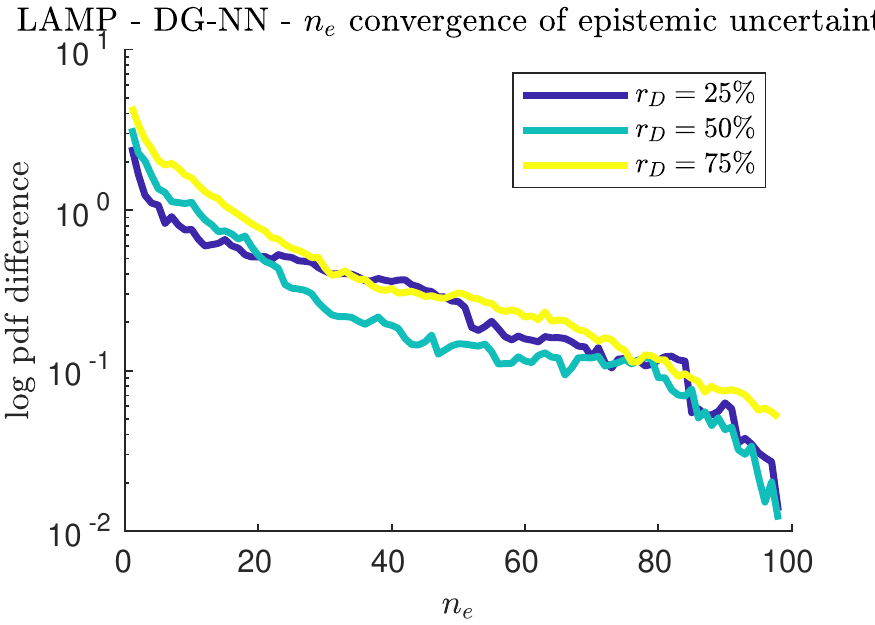}
\includegraphics[width=0.45\linewidth, trim=0 0 0 0, clip] {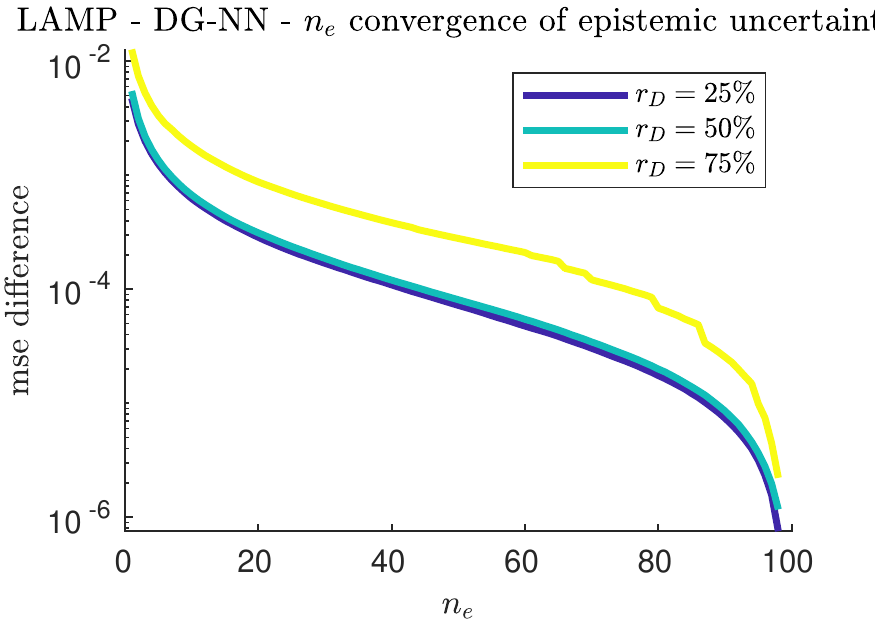}
\caption{Convergence curves of $\sigma_\epsilon$ for different $r_D$ as functions of ensemble size, $n_e$. \textbf{Left}: log pdf difference: $\int{| \log{p_S(\sigma_\epsilon^{n_e})} - \log{p_S(\sigma_\epsilon^{100})} |\text{d}\sigma_\epsilon}$.  \textbf{Right}: Mean of squared differences: $\sum{(\sigma_\epsilon^{n_e} - \sigma_\epsilon^{100}})^2/N_{samples}$.}
\label{fig:dnn-ens_conv}
\end{figure}

In figure \ref{fig:dnn-nd}, we display some of the effects of $r_D$ and $n_e$ on the normalized residuals and epistemic uncertainty distributions.  In the left subfigure, we show how changing $r_D$ and $n_e$ affects the distribution of uncertainties.  On the right, we see how changing $r_D$ and $n_e$ affects the distribution of normalized residuals.  We note that increasing $r_D$ has complicated effects on the distribution of $\sigma_\epsilon$, but it generally increases $\sigma_n$ and the range of $\sigma_\epsilon$.  Additionally, we note that increasing $r_D$ causes the distribution of normalized residuals to more closely approximate the standard Gaussian. Figure \ref{fig:dnn-nd} implies that increasing $n_e$ from 5 to 50 does not have any significant effects on normalized residuals and epistemic uncertainty distribution. Additionally, figure \ref{fig:dnn-ens_conv} depicts the effects of ensemble size, $n_e$, and $r_D$ on estimating $\sigma_\epsilon$ both in value and distribution.

\noindent\textbf{Parameter Study.}
In the previous section, we hinted at one of the parameters that directly affect the performance of Dropout Neural Networks (D-NNs) and Dropout Gaussian Neural Networks (DG-NNs), i.e., ensemble size ($n_e$). We showed the convergence of $\log{p_S(\sigma_\epsilon^{n_e})}$ and the mean squared error in $\sigma_\epsilon$ vs. ensemble size ($n_e$) - cf. figure \ref{fig:dnn-ens_conv}. While it shows orders of magnitudes improvement as $n_e$ increases, figure \ref{fig:dnn-nd} shows that qualitatively, even an ensemble size of $n_e = 5$ can accurately predict unseen data and estimate epistemic uncertainties compared to $n_e = 50$. 

Additionally, figure \ref{fig:dnn-ens_conv} shows the effects of another parameter, dropout rate ($r_D$), on DG-NN prediction of unseen data (implied in distribution of normalized residuals) and estimation of epistemic and aleatoric uncertainties. We can make the following conclusions, in terms of bias-variance trade-off, by examining figure \ref{fig:dnn-nd}:
\begin{itemize}
    \item \textbf{Small dropout rate} ($r_D = 25\%$): The model in the case with small dropout rate (blue curves) estimates low uncertainty values (lower aleatoric uncertainty and epistemic uncertainty mode). In addition, the large tails of the normalized residuals distribution (compared to a normal distribution) corresponding to $r_D = 25\%$ implies that the model is unable to generalized to previously not seen data. In summary, with small dropout rate the model tends to overfit, i.e., \textbf{low bias - high variance}. 
    
    \item \textbf{Large dropout rate} ($r_D = 75\%$): In the case of large dropout rate (yellow curves), the model estimates much larger uncertainty values (larger aleatoric uncertainty and epistemic uncertainty mode). Also, the narrow nature of the associated normalized residuals distribution (narrow compared to a normal distribution) implies that the model generalizes \textit{too well} to unseen data. The conclusion here is that large dropout rates create models that tend to underfit, i.e., \textbf{high bias - low variance}. 
    
    \item \textbf{50$\%$ dropout rate} ($r_D = 50\%$): We have observed that the model with $r_D = 50\%$ not only estimates the uncertainties well (compared to the other two dropout rates) but also generalizes well to unseen data as evident by its corresponding normalized residuals distribution. Our conclusion in this case is that a DG-NN model with 50$\%$ dropout rate properly trains on the training data and generalizes well to unseen data, as well, i.e., \textbf{medium bias - medium variance}. 
    
\end{itemize}

\subsection{Functional Inputs}
\label{sec:neural-op}

\begin{figure}[htpb]
\centering
\begin{tikzpicture}[node distance = 0.5cm, thick,
roundnode/.style={circle, draw=green!60, fill=green!5, very thick, minimum size=12mm},
squarednode/.style={rectangle, draw=red!60, fill=red!5, very thick, minimum size=8mm}
]%
		\node[roundnode, align=center] (x0) {$x(t)$};
		\node[squarednode, align=center] (x1) [right=of x0] {Reduced \\ Order \\ Model};
        \node[roundnode, align=center] (x2) [right=of x1] {$\vec{\alpha}$};
		\node[squarednode, align=center] (x3) [right=of x2] {ROM \\ Neural \\ Network};
        \node[roundnode, align=center] (x4) [right=of x3] {$y$};
        \node[squarednode, align=center] (y0) [below=of x2] {Functional Neural Network};

        \draw[->] (x0.east) -- (x1.west);
        \draw[->] (x1.east) -- (x2.west);
        \draw[->] (x2.east) -- (x3.west);
        \draw[->] (x3.east) -- (x4.west);

        \draw[->] (x0.south) to [out=270,in=180] (y0.west);
        \draw[->] (y0.east) to [out=0,in=270] (x4.south);

\end{tikzpicture}%
\caption{Flowchart model for functional input.}
\label{fig:functional-input}
\end{figure}
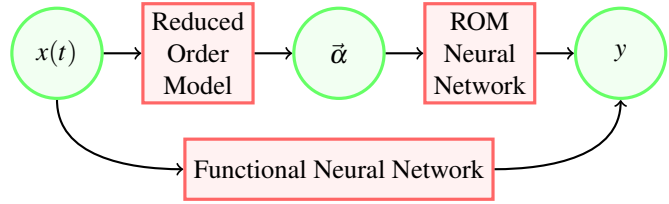

So far, we have considered the input to our machine learning models to be coefficient vectors for a reduced order model (ROM):  wave time series for the LAMP example and initial condition for the MMT example.  This choice is appropriate for GP, which struggle with higher dimensional input.  However, neural network architectures are not so limited.

We may consider instead feeding a time series or spatial series input into a first fully connected layer of the neural network, after which the network architecture would be the same as before.  In the LAMP case, this first layer would be a 1024 dimensional vector, representing the time series sampled uniformly on the interval $[0, T]$, where $T=40$ is the duration of the wave episodes considered in the LAMP simulation.  In the MMT case, this first layer would also be a 1024 dimensional vector, the first 512 components of which correspond to the real part of the PDE initial conditions on the interval $[0, 1]$, and the second 512 components of which correspond to the imaginary parts.

In figure \ref{fig:functional-input}, we show a cartoon relationship between the the neural network described previously, the ROM network, and the functional network.  In the ROM network, we include a preprocessing step to transform the input into a low dimensional vector representation.  Conversely, in the functional network, we feed the input directly into a fully connected layer of the network.  In each case, the network returns a scalar output.  This may be compared to the literature on neural operators such as DeepONet and Fourier Neural Operator \cite{bhattacharya21, kovachki21,lu22}, where both the input and the output side of the network are functions.

We make a special note that we keep the same network architecture--we have not introduced recurrent networks or convolutional networks.  The functional input mode is compatible with all of the UQ techniques discussed above, but takes little advantage of the most recent developments in neural operator architecture.  In section \ref{sec:res-n-o}, we present some comparisons of surrogate models using ROM input and functional input.

\subsection{Combinations and Summary}
In figure \ref{tab:summary-model-uq}, we summarize the uncertainty quantification ability of each machine learning technique discussed.  In general, the G-NN final layer (included in BNN) is required to estimate aleatoric uncertainty.  Further, some kind of ensemble technique is required to estimate epistemic uncertainty.  However, both BNN and DG-NN provide an alternative ensemble technique, and DG-NN generally overperformed compared to ENN.

\begin{table}[htpb]
\centering
\begin{tabular}{ c c c c }
 Architecture & Epistemic UQ? & Aleatoric UQ? \\
 \hline  \hline
 GP & yes & yes \\ 
 NN & no & no \\
 G-NN & no & yes \\ 
\hline
 ENN & yes & no \\ 
 D-NN & yes & no \\
  \hline
 BNN & yes & yes \\
 EG-NN & yes & yes \\ 
 DG-NN & yes & yes
\end{tabular}
\caption{Comparison of UQ capabilities for various ML techniques.}
\label{tab:summary-model-uq}
\end{table}

Each of the large scale architectures we trialed also has a number of adjustable hyperparameters.  We summarize our final selections in tables \ref{tab:hyperparameters-lamp} and \ref{tab:hyperparameters-mmt}. We note that the width and depth of our neural networks, $8\times[250]$, is far into the overparameterization regime.  Our choice of activation function, ReLU, is also a standard activation function.  Some techniques, such as BNN and EG-NN were found to be sensitive to choice of activation function, while others, like DG-NN, were not.
\begin{table}[htpb]
\centering
\begin{tabular}{ c c c c c }
 Architecture & Depth & Activation Function & loss function & explicit regularization \\
 \hline  \hline
 ENN & $8\times[250]$ & ReLU & mse & yes \\  
 EG-NN & $8\times[250]$ & ReLU & NLL & yes \\  
 BNN & $8\times[250]$ & ReLU & -ELBo & no \\
 DG-NN & $8\times[250]$ & ReLU & NLL & no 
\end{tabular}
\caption{Selection of architecture details used in neural network models for LAMP presented below.}
\label{tab:hyperparameters-lamp}
\end{table}

\begin{table}[htpb]
\centering
\begin{tabular}{ c c c c c }
 Architecture & Depth & Activation Function & loss function & explicit regularization \\
 \hline  \hline
 ENN & $8\times[250]$ & ReLU & mse & yes \\  
 D-NN & $8\times[250]$ & ReLU & mse & no 
\end{tabular}
\caption{Selection of architecture details used in neural network models for MMT presented below.}
\label{tab:hyperparameters-mmt}
\end{table}

Additionally, we note that Dropout, and to a lesser extent the Variational Inference technique behind BNN, are implicit regularization strategies.  For ENN and EG-NN, which cannot take advantage of this built in regularization, we used $L_2$ regularization of weights.  

\section{Quality Measures}
\label{sec:metrics}

To measure the quality of each regression method we cannot simply rely on quantifying their regression accuracy, as this does not contain information for the uncertainty quantification part. To this end we adopt two separate measures. 

\subsection{Normalized Residual Distribution}

The first quality metric we examine is the distribution of \textit{normalized residuals} (NR), sometimes called z-scores. This is given by the histogram of the normalized error \textit{computed on the validation set}:

\begin{equation}
\label{eq:norm-res}
    z_i = \frac{y_i - \mu(\alpha_i)}{\sigma(\alpha_i)}.
\end{equation}

The NR histograms measures the self-consistency of the regression scheme, i.e. how honest it is on accurately predicting the regions of the input space where the prediction uncertainty is high. It is not directly related to the quality of the regression, i.e. how accurate it is, but rather to whether the errors on the validation data are consistent in magnitude and location with the a priori estimated uncertainty. For accurately estimated uncertainties, the NR histogram should match a distribution with zero mean and unit variance.  

When we present NR plots, we include an overlaid standard Gaussian distribution ($\mathcal{N}(0, 1)$) to help with evaluation.  We emphasize that a well calibrated UQ scheme will not, in general, match the \textit{shape} of the standard Gaussian distribution, but will match the \textit{mean} (i.e., $0$) and the \textit{variance} (i.e., $1$). Generally, a NR histogram with variance greater than one, or generally heavier tails, is an indicator of over-fitting for the corresponding regression scheme, i.e. larger errors than what is expected by the estimated uncertainty. On the other hand, if the NR variance is less than one, we have higher estimated uncertainty than the actual errors indicate. This could be the case, for example, when the regression schemes operates as a smoother over the data, a behavior typically combined by a large uncertainty (aleatoric or epistemic).  Alternately, this can occur from unconverged model fitting.

For each of our test cases, we generate the distribution of normalized residuals by separating the data into a training set and a validation set.  For each point in the validation set, none of which have been seen by the model during training, we compute the residual (model error) and divide by the total model uncertainty at that point.  We emphasize that the set of normalized residuals computed this way are \textbf{not} a distribution, because the samples from the validation set are are not independent of the samples from the training set--in particular, the validation data are precisely the samples not selected for the training set!  This proviso is more significant for the LAMP data, where computational costs limit the total number of simulations, than for the MMT data, where only a small fraction of the total data is used for training.

\subsection{Uncertainty Distribution}

While NR quantifies the consistency of the estimated uncertainty with the model errors, it does not provide an overall measure of the accuracy of the regression method. To cover this aspect, we also adopt the uncertainty histogram, i.e. a histogram of all the estimated epistemic uncertainties, $\sigma(\alpha_i)$, over the validation set. For models that include an estimate of aleatoric uncertainty, we additionally include the estimated aleatoric uncertainty--for this work, assumed to be uniform over the input space. It is important to emphasize that the distribution of epistemic uncertainties and the aleatoric uncertainty measure how accurate each regression scheme `believes' it is, through its own uncertainty quantification scheme.

Whether this uncertainty is consistent with the accuracy of the regression, it is measured by the NR distribution. To this end, these two measures should be used in combination to quantify the quality of any regression schemes that aims to quantify uncertainty.  A high quality regression scheme is associated with small values of the uncertainty, and an NR distribution with unit variance.



\section{Results}
\label{sec:results}

\subsection{LAMP Dataset}

\begin{figure}[]
\centering
\includegraphics[width=0.45\linewidth, trim=0 0 0 0, clip] {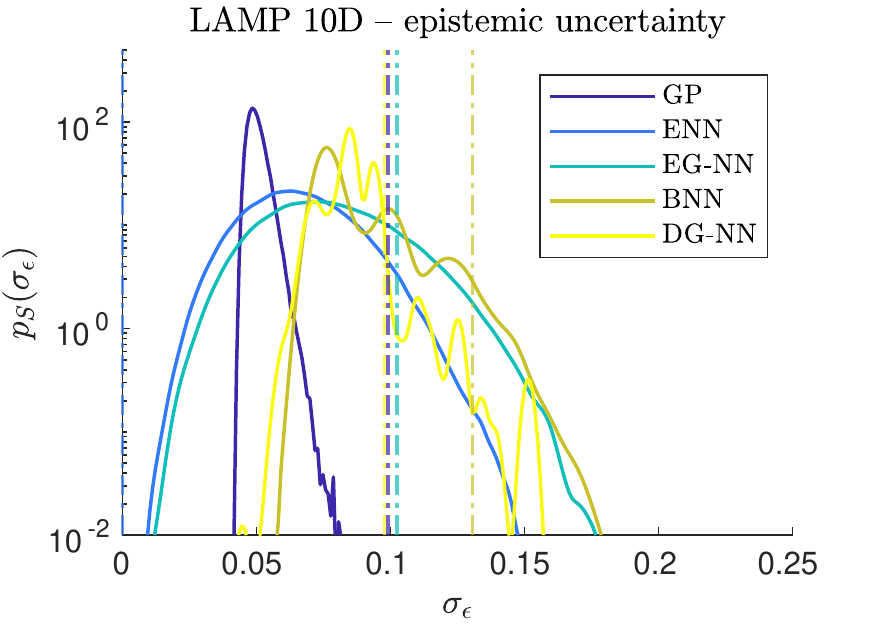}
\includegraphics[width=0.45\linewidth, trim=0 0 0 0, clip] {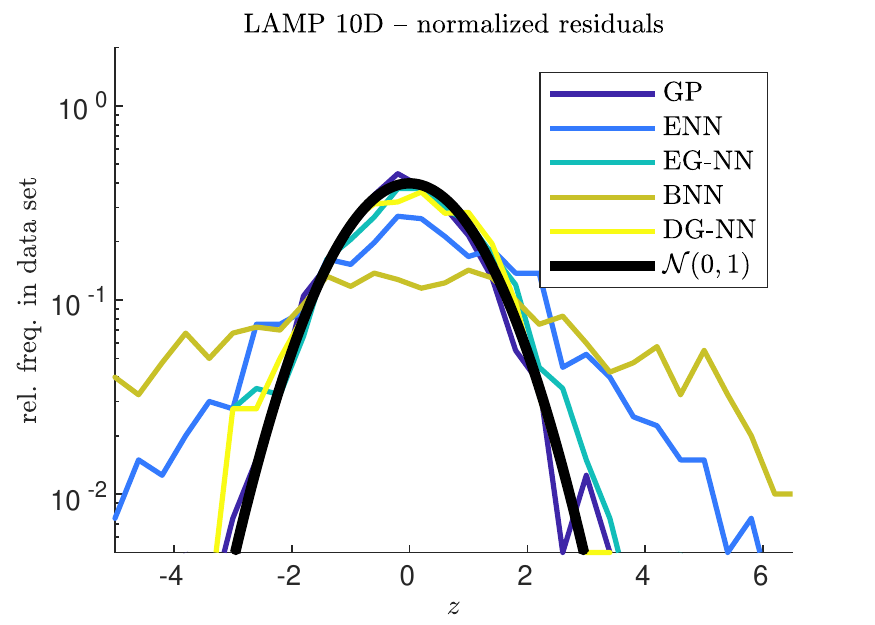}
\caption{\textbf{Left}:  Distributions of epistemic uncertainty $\sigma_\epsilon$, for LAMP data and different ML techniques.  $N=10, n_s=2000$.  \textbf{Right}:    Distributions (pdf) of normalized residuals $z_i = \frac{y_i - \mu(\alpha_i)}{\sigma(\alpha_i)}$.}
\label{fig:comp-epi-norm-res-pdf-10d}
\end{figure}

First, we consider UQ for the LAMP dataset.  In figure \ref{fig:comp-epi-norm-res-pdf-10d}, we display the distribution of uncertainties, as well as the normalized residuals for $N=10$ dimensional LAMP data.  In the left column, we display the pdf of epistemic uncertainties for each ML technique, along with a vertical line for the estimate aleatoric uncertainty, $\sigma_n$.   
On the right column, we plot the normalized residuals, given by equation \ref{eq:norm-res}.  We note that GP, EG-NN, and D-GNN all estimate a very similar value for $\sigma_n$. It is important to emphasize that despite the relatively high dimensionality of the data set its inherently low dimensional structure allows for GP to work very well in terms of minimizing epistemic uncertainty. The NN based methods have comparable performance in terms of epistemic uncertainty. In terms of UQ properties, i.e. capability to model own errors,  besides ENN and BNN both of which underestimate model error, the UQ for GP, EG-NN, and DG-NN are all well calibrated.

\subsection{MMT Dataset}

\begin{figure}[]
\centering
\includegraphics[width=0.45\linewidth, trim=0 0 0 0, clip] {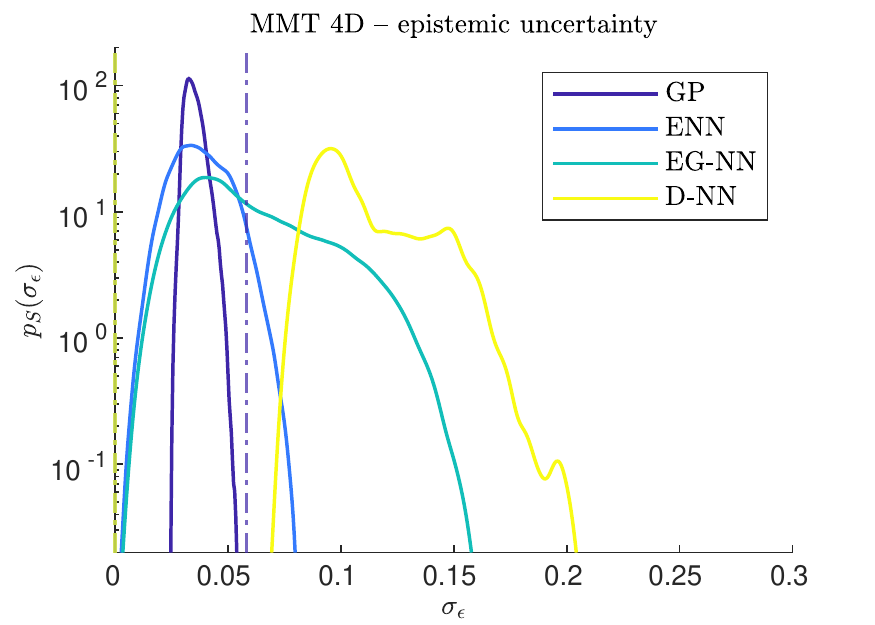}
\includegraphics[width=0.45\linewidth, trim=0 0 0 0, clip] {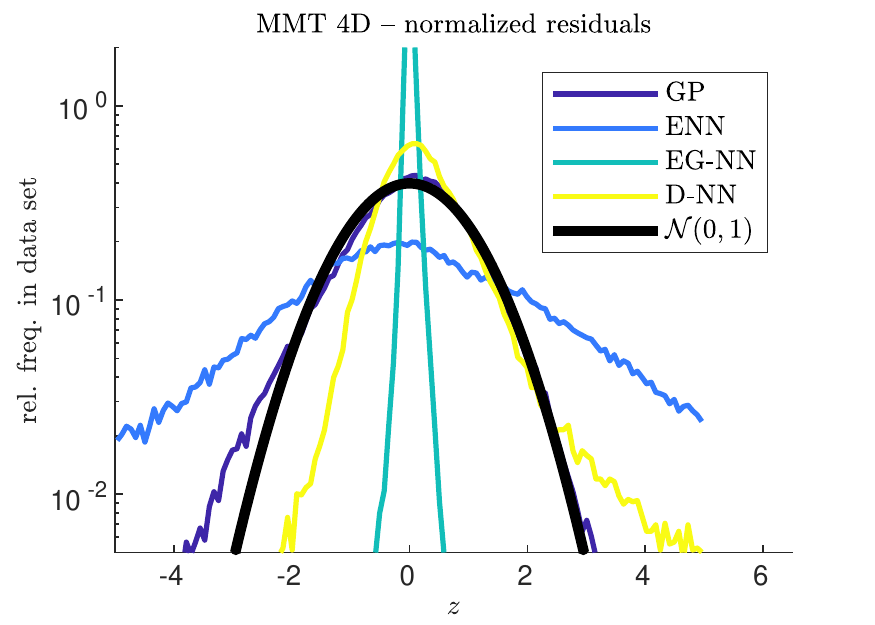}
\caption{\textbf{Left}:  Distributions of epistemic uncertainty $\sigma_\epsilon$, for MMT data and different ML techniques.  $N=4, n_s=2000$.  \textbf{Right}:    Distributions of normalized residuals $z_i = \frac{y_i - \mu(\alpha_i)}{\sigma(\alpha_i)}$.}
\label{fig:mmt-comp-epi-norm-res-pdf-4d}
\end{figure}

\begin{figure}[]
\centering
\includegraphics[width=0.45\linewidth, trim=0 0 0 0, clip] {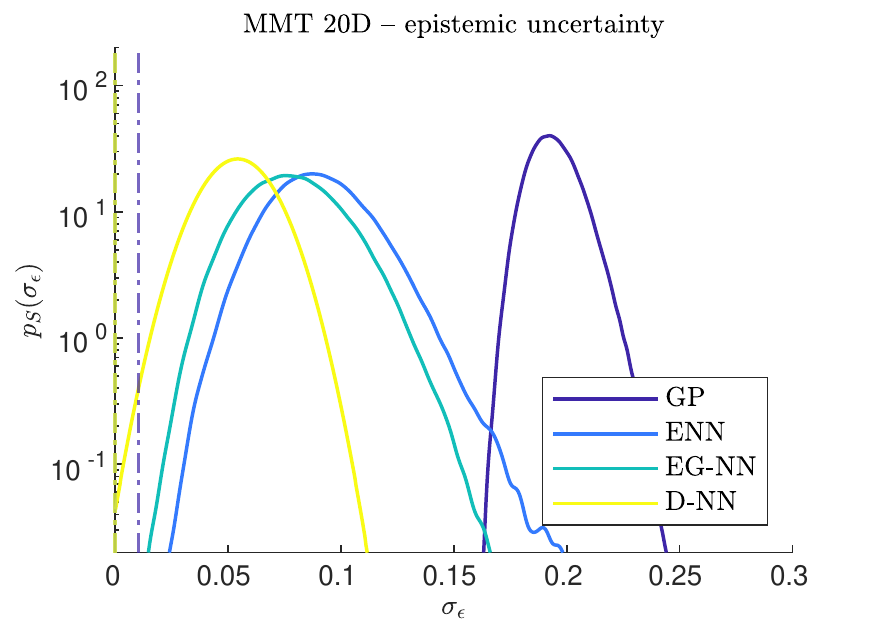}
\includegraphics[width=0.45\linewidth, trim=0 0 0 0, clip] {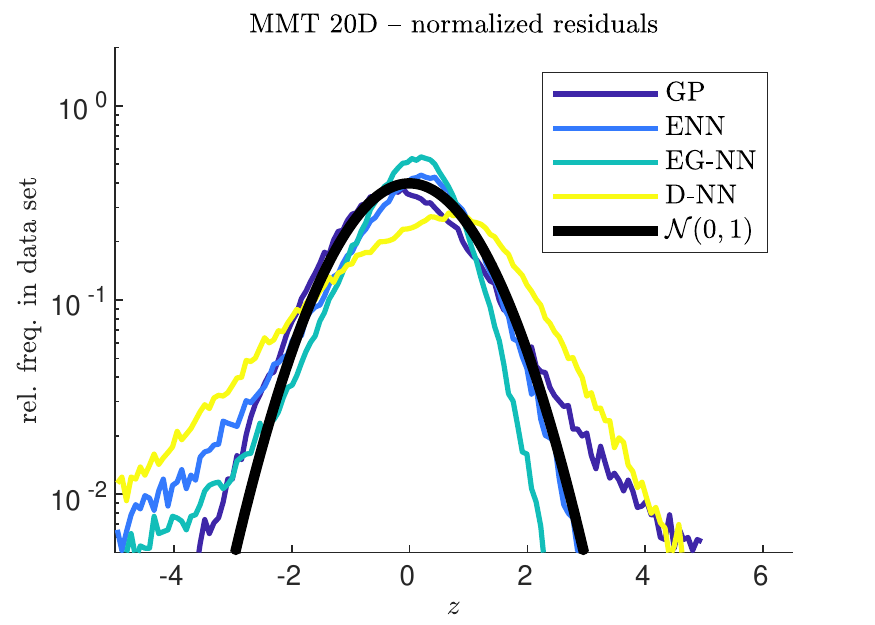}
\caption{\textbf{Left}:  Distributions of epistemic uncertainty $\sigma_\epsilon$, for MMT data and different ML techniques.  $N=20, n_s=10000$.  \textbf{Right}:    Distributions of normalized residuals $z_i = \frac{y_i - \mu(\alpha_i)}{\sigma(\alpha_i)}$.}
\label{fig:mmt-comp-epi-norm-res-pdf-20d}
\end{figure}

In figures \ref{fig:mmt-comp-epi-norm-res-pdf-4d} and \ref{fig:mmt-comp-epi-norm-res-pdf-20d}, we compare the UQ performance for selected models on MMT datasets of dimensionality $N=4$ and $N=20$.  We chose not to include BNN in this comparison, because it performed poorly in the LAMP trial and preliminary investigation for MMT confirmed this performance. We further note that we include D-NN instead of DG-NN for this dataset.  This choice was made to capitalize on our a priori knowledge that the MMT dataset does not include aleatoric uncertainty, combined with our experience that the G-NN final layer complicates training when the true $\sigma_n \approx 0$.

We observe that GP performs fairly well compared to NN architectures for $N=4$ dimensions in terms of estimated uncertainty, but performs its performance deteriorates for  $N=20$.  This is in line with the intuition that GP scales poorly to higher dimensional input space.  Note that while the GP model uncertainty/errors are large absolutely in $N=20$ dimensions, the UQ is still well calibrated according to the normalized residual plot. On the other hand, for the high dimensional case, the NN based architectures perform very well in minimizing epistemic uncertainties with the D-NN having the best performance. Moreover, they have good UQ-error consistency properties with the D-NN doing slightly worse compared with the other methods, i.e. its UQ underestimates its large errors.

\subsection{Functional Inputs}
\label{sec:res-n-o}

\begin{figure}[htpb]
\centering
\includegraphics[width=0.45\linewidth, trim=0 0 0 0, clip] {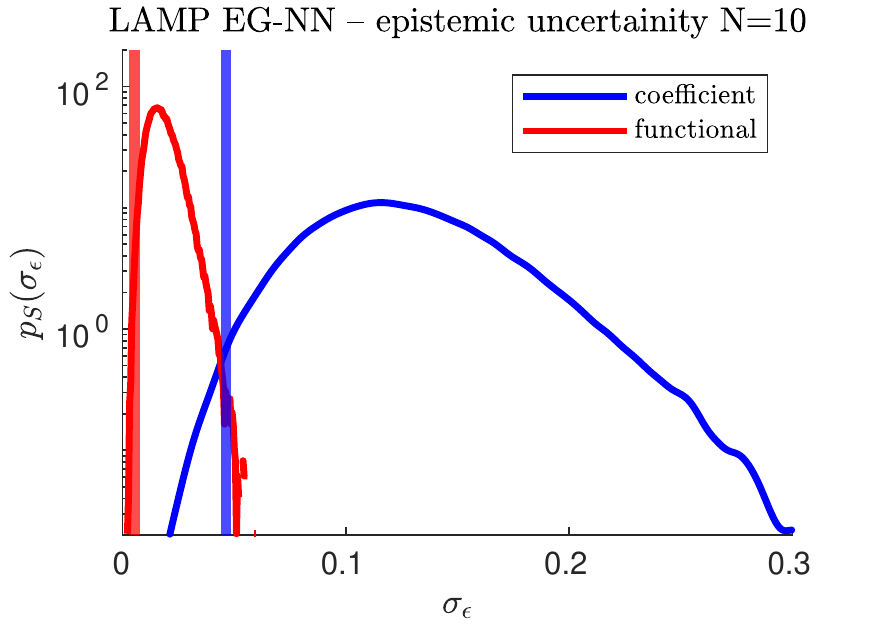}
\includegraphics[width=0.45\linewidth, trim=0 0 0 0, clip] {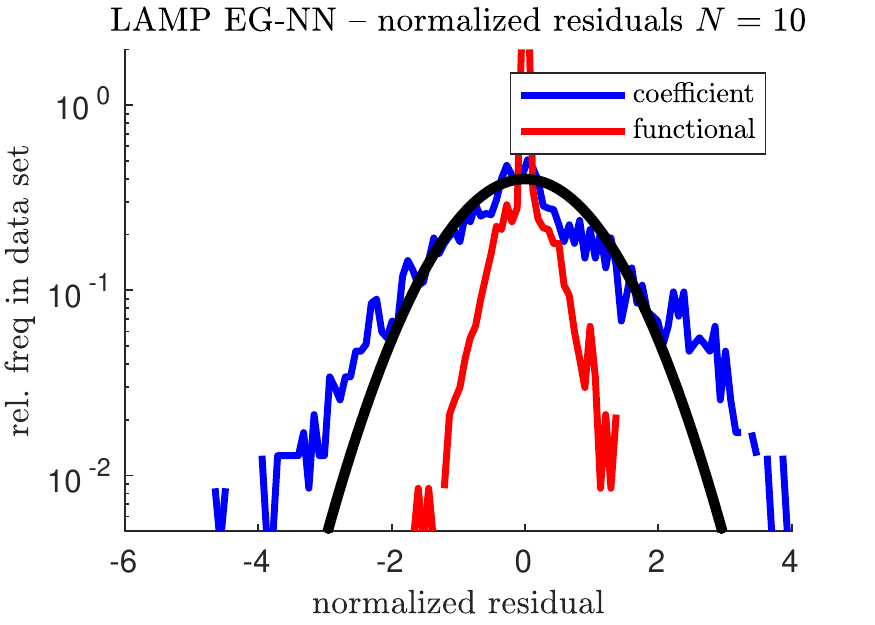}
\caption{Comparison of EG-NN with ROM inputs and with functional inputs for 10D LAMP data, $n_s=2000$.  \textbf{Left}:  Comparison of uncertainties.  \textbf{Right}:  Comparison of normalized residuals.}
\label{fig:rom-vs-func-lamp-10d}
\end{figure}

\begin{figure}[htpb]
\centering
\includegraphics[width=0.45\linewidth, trim=0 0 0 0, clip] {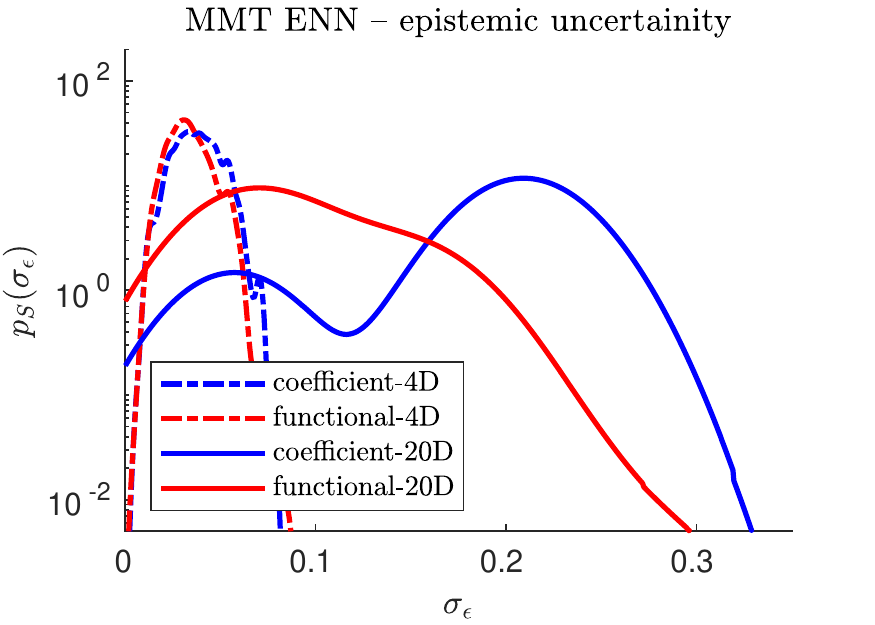}
\includegraphics[width=0.45\linewidth, trim=0 0 0 0, clip] {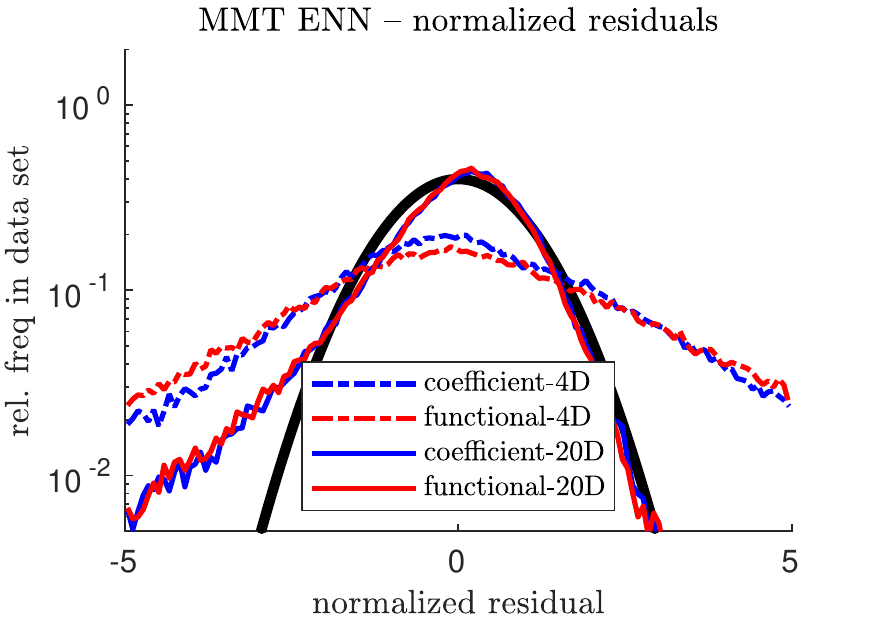}
\caption{Comparison of ENN with ROM inputs and with functional inputs for 4D and 20D MMT data ($n_s=462$ and $n_s=10000$).  \textbf{Left}:  Comparison of uncertainties.  \textbf{Right}:  Comparison of normalized residuals.}
\label{fig:rom-vs-func-mmt}
\end{figure}

In addition to comparisons between different surrogate models, we also compared the ENN and EG-NN techniques using ROM inputs and functional inputs as described above in section \ref{sec:neural-op}.  In figures \ref{fig:rom-vs-func-lamp-10d} and \ref{fig:rom-vs-func-mmt}, we present comparisons for for both LAMP and MMT data sets. In figure \ref{fig:rom-vs-func-lamp-10d}, we see that functional inputs lead to significantly better predictions in the LAMP problem with a 10 dimensional space of wave episodes.  This is evidenced by a distribution of estimated model errors (left plot) much smaller for functional inputs (red), without concomittant overfitting (right plot).  In fact, the observed \textit{underfitting} suggests that network expressitivity is a limiting factor. 

In figure \ref{fig:rom-vs-func-mmt}, we see that for the MMT data, the functional inputs lead to smaller improvements.  While for 4D input data, the different input methods has no clear difference, but 20D input data, the functional input mode shows a clear improvement, if not as dramatic as displayed by the LAMP data. Why does functional input improve the results from LAMP data more than for MMT data? It seems that functional neural networks have a stronger advantage when the underlying structure of the data set has lower dimensionality (section \ref{sec:compare-data-sets}). Thus, the LAMP wave episodes, which inherit more structure from their KL basis, are able to better take advantage of the functional input improvement, compared with the MMT data set.

\section{Conclusions}

In this paper, we have reviewed a number of machine learning methods for combining neural network surrogate models with UQ for estimating both aleatoric and epistemic uncertainties in data sets associated to complex dynamical systems.  First, we presented the G-NN model, which allows any neural network architecture, including traditional deterministic feed-forward networks, to estimate aleatoric uncertainty associated with a training data set.  Next, we presented three implementations of the ensemble technique for for estimating epistemic uncertainty:  ENN, BNN and D-NN.  Each of these ensemble techniques may be combined with the G-NN final layer, allowing for a neural network surrogate model able to estimate both aleatoric and epistemic uncertainties.  Finally, motivated by recent work on neural operators, we described how to simply change the network architecture to handle functional inputs.

We compared each of these techniques, along with a GP model for reference, on two different real datasets obtained by complex dynamical systems:  one with an important intrinsic aleatoric component (LAMP) and the other without (MMT).  We compared performance with metrics tailored specifically to the question of UQ:  normalized residuals and predicted uncertainties.  The distribution of normalized residuals from a hold-out test set provide an estimate of how well calibrated the model UQ is and may indicated presence of underfitting or overfitting.  The distribution of the model uncertainties indicated how uncertain the model is generally.

In the data set with intrinsically low dimensional structure (LAMP) as well as the low-dimensional version of the MMT data set, GP always performs the best: it has one of the the lowest epistemic uncertainties, as it is also self consistent, i.e. errors in the validation test are correlated well with the estimated uncertainty in terms of magnitude and location. On the other hand, for the high dimensional data set MMT, GP has important limitations, as expected. In this case, ensemble methods, in particular D-NN or ENN perform consistently better. They also have excellent performance in the normalized residual test. We found that BNN models underperformed, especially compared to D-NN, their most similar competitor.  Further, there were also significant difficulties in training BNN models. Regarding the estimation of the aleatoric uncertainty, we found that adding a Gaussian layer in ensemble methods (EG-ENN, DG-NN) results in accurate estimation of the aleatoric uncerainty, i.e. comparable with GP.

Finally, we examined the use of ROM (coefficient) inputs compared to functional inputs.  In the case of the LAMP data, functional inputs performed significantly better than coefficient inputs with minimal tuning.  However, in the case of MMT data functional inputs provided only marginal improvements after significant optimization.  We conjecture that this difference is due to differences in the inherent dimensionality of the two data sets.

Overall, the GP is the recommended method for data sets that `live' in low dimensional spaces or have a low-dimensional structure. For intrinsically high-dimensional data sets we recommend the dropout technique for constructing ensemble variance estimates, with a balanced dropout fraction of $r_D=0.5$.  However, we find that the ensemble networks may also be useful, especially in situations where complicated model architecture may make the dropout technique challenging to implement.

\subsection*{Acknowledgments} We acknowledge support from the AFOSR MURI grant no. FA9550-21-1-0058, the DARPA grant no. HR00112290029, and the ONR grant no. N00014-21-1-2357. We are grateful to Dr. Ethan Pickering for useful discussions.

\section*{Appendix}
\label{sec:appendix}

\subsection{LAMP Data N=2 -- Visualization of Surrogate and Epistemic Uncertainty}
\label{sec:appendix-lamp}

\begin{figure}[htpb]
\centering
\includegraphics[width=0.4\linewidth, trim=0 0 0 0, clip] {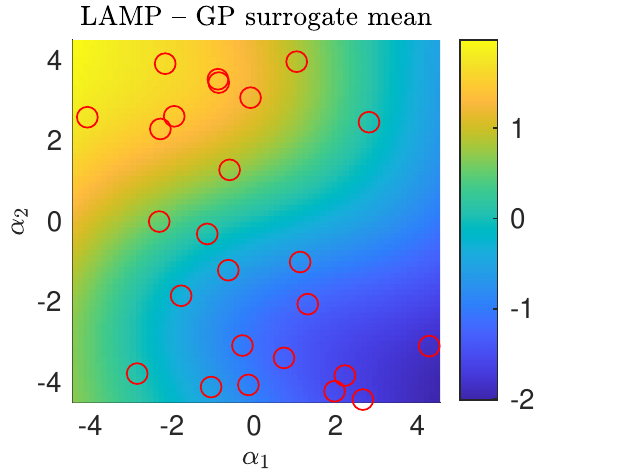}
\includegraphics[width=0.4\linewidth, trim=0 0 0 0, clip] {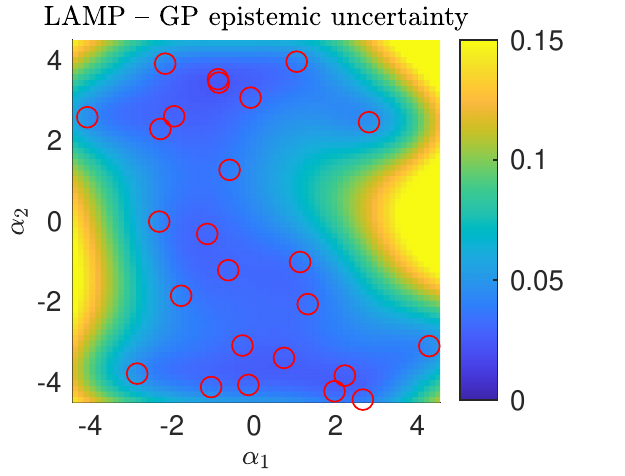}
\includegraphics[width=0.4\linewidth, trim=0 0 0 0, clip] {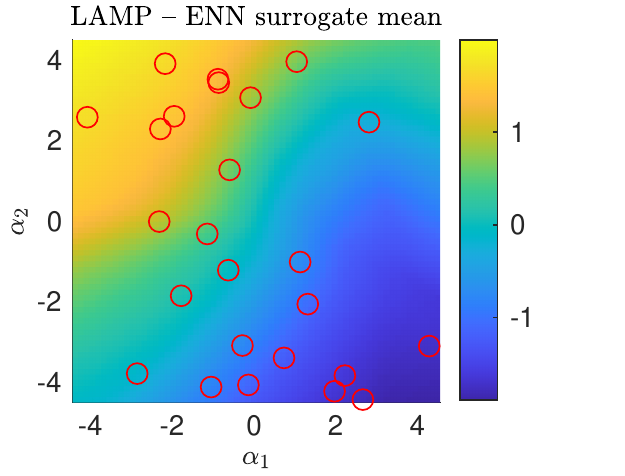}
\includegraphics[width=0.4\linewidth, trim=0 0 0 0, clip] {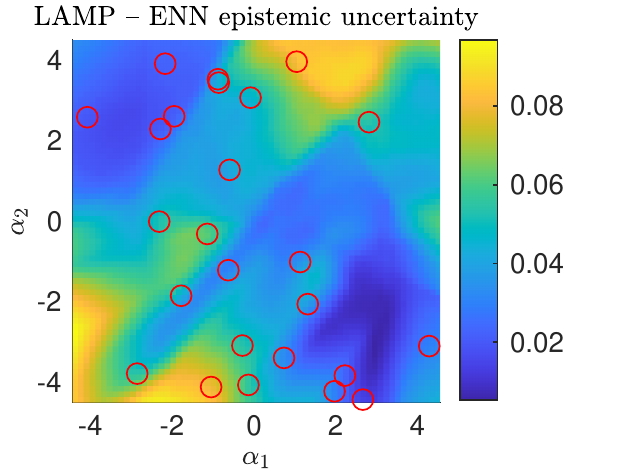}
\includegraphics[width=0.4\linewidth, trim=0 0 0 0, clip] {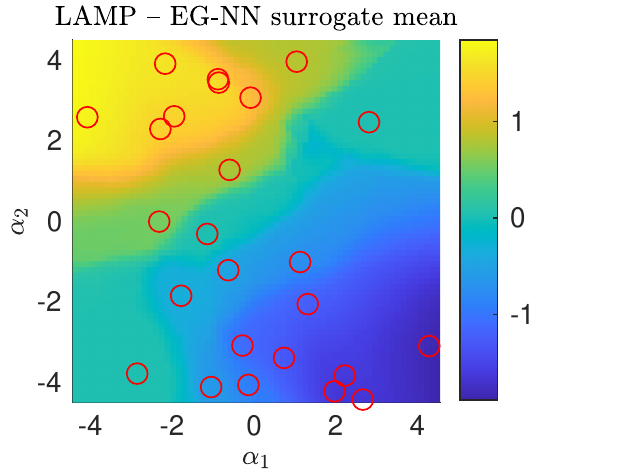}
\includegraphics[width=0.4\linewidth, trim=0 0 0 0, clip] {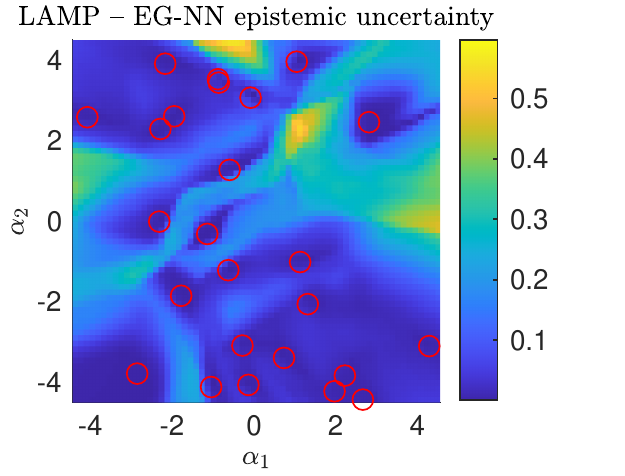}
\includegraphics[width=0.4\linewidth, trim=0 0 0 0, clip] {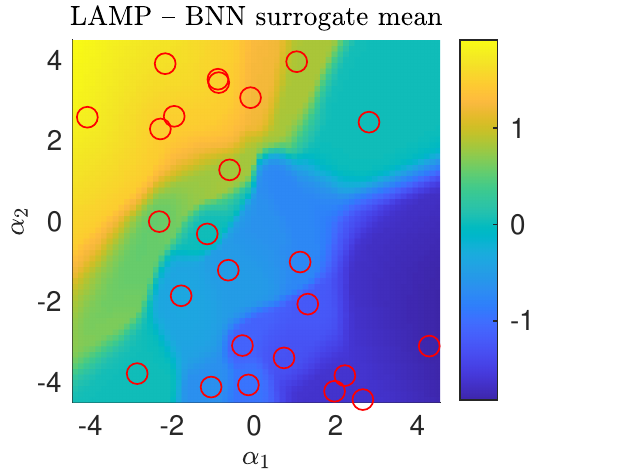}
\includegraphics[width=0.4\linewidth, trim=0 0 0 0, clip] {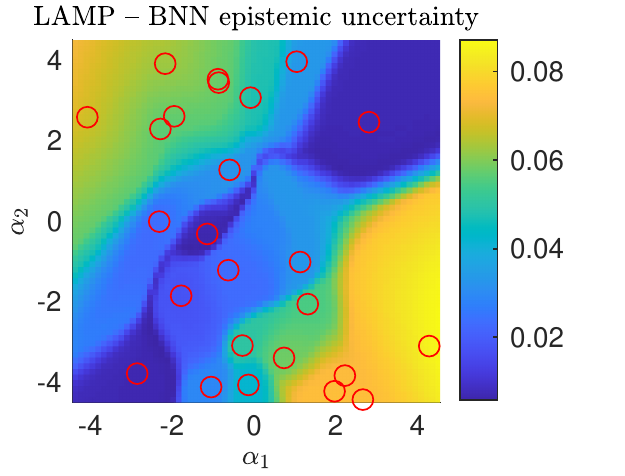}
\includegraphics[width=0.4\linewidth, trim=0 0 0 0, clip] {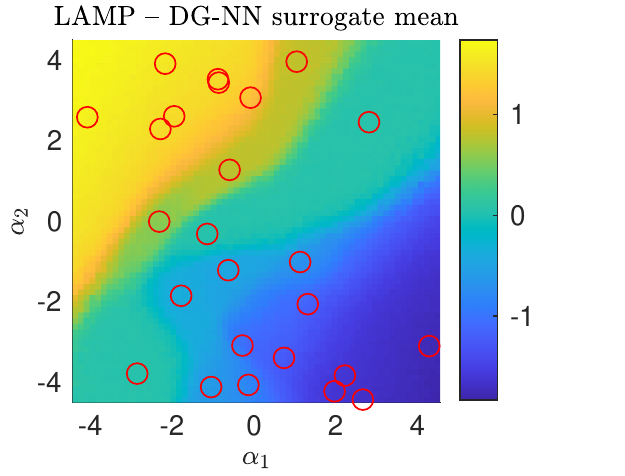}
\includegraphics[width=0.4\linewidth, trim=0 0 0 0, clip] {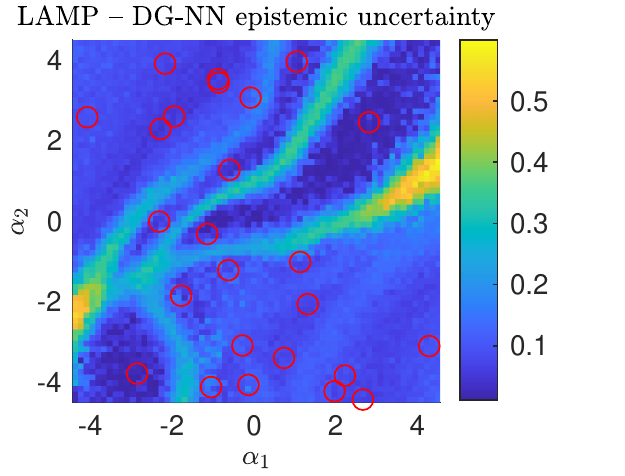}
\caption{Spatial distributions of the (\textbf{Left}) surrogate mean $\mu(\vec{\alpha})$ and (\textbf{Right}) epistemic uncertainty $\sigma_\epsilon(\vec{\alpha})$, for LAMP data and different ML techniques.  $N=2, n_s=25$.}
\label{fig:comp-2d-surrogate}
\end{figure}

A reasonable question for comparison across these techniques is, where is the epistemic uncertainty maximized?  This is important for uncertainty quantification generally, but especially for active learning with uncertainty sampling based techniques.

In figure \ref{fig:comp-2d-surrogate}, left column, we show the surrogate mean $\mu(\vec{\alpha})$ in the restricted $N=2$ dimensional case, with $n_s=25$ training points (c.f. figure \ref{fig:data-surr-models}).  We show this image for background before we display the spatial distribution of epistemic uncertainties in the right column of figure \ref{fig:comp-2d-surrogate}.  In particular, note the belt of maximum gradient passing along an approximately 45 degree angle near the origin. In figure \ref{fig:comp-2d-surrogate}, right column, we show epistemic uncertainty as a function of $(\alpha_1, \alpha_2)$ for a 2D dataset with $n_s=25$.  The low dimension and small training set allows one to compare the location of the training data to the location of regions of high epistemic uncertainty.

We note that, while generally we hold up GP as the gold standard of uncertainty quantification, for the question of the spatial distribution of epistemic uncertainty we give GP no special deference.  The kernel matrix math for $\sigma(\vec{\alpha})$ is independent of the $q_1$ component of the training data.  As a result, GP uncertainty estimation largely ignores the existence of regions with large gradient, except to the extent they are captured during the hyperparameter optimization.

Additionally, the spatial distribution of the epistemic uncertainties (and not the absolute values) is of particular important to active sampling.  In uncertainty sampling based active sampling methods, the activation function is a product of the epistemic variance, possibly along with some weight factor:

\begin{equation}
    u_{\mbox{US}}(\vec{\alpha}) = w(\alpha)\sigma^2_{\epsilon}(\vec{\alpha}).
\end{equation}

\noindent In this application, scaling all the uncertainties by a constant factor leaves the relative ordering of the acquisition function unchanged.

\subsection{MMT Data N=2 -- Visualization of Surrogate and Epistemic Uncertainty}

\begin{figure}[htpb]
\centering
\includegraphics[width=0.4\linewidth, trim=0 0 0 0, clip] {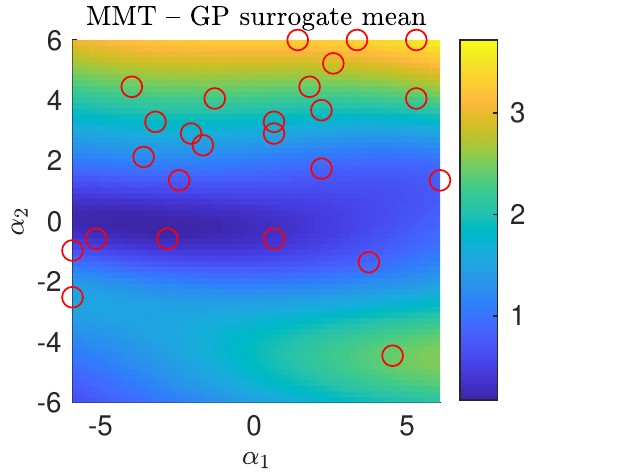}
\includegraphics[width=0.4\linewidth, trim=0 0 0 0, clip] {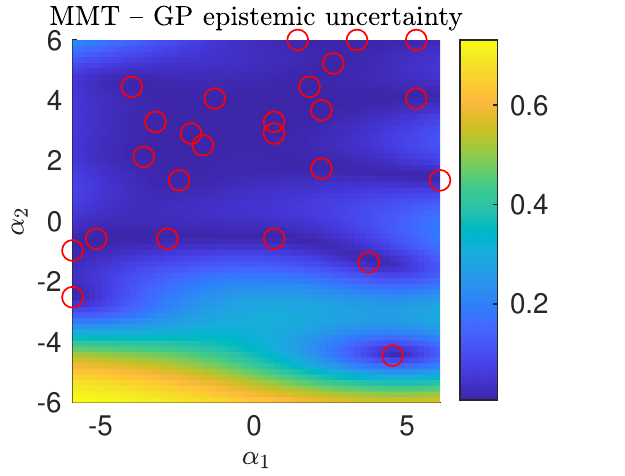}
\includegraphics[width=0.4\linewidth, trim=0 0 0 0, clip] {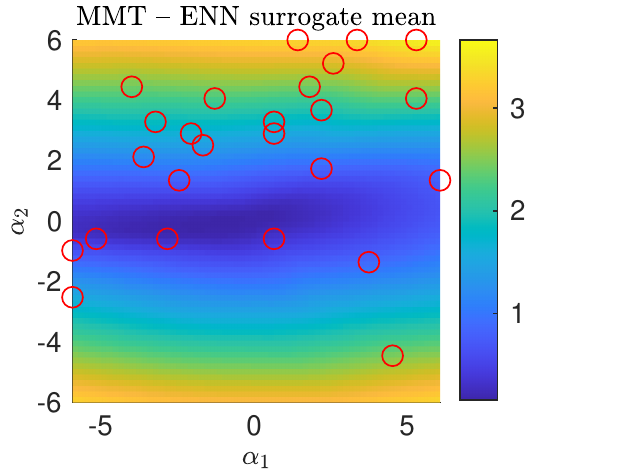}
\includegraphics[width=0.4\linewidth, trim=0 0 0 0, clip] {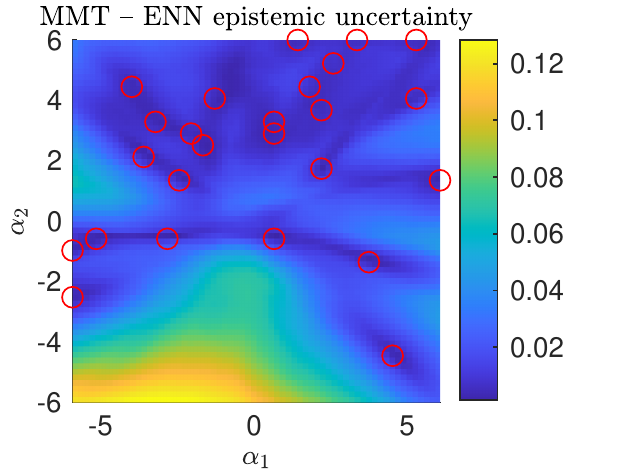}
\includegraphics[width=0.4\linewidth, trim=0 0 0 0, clip] {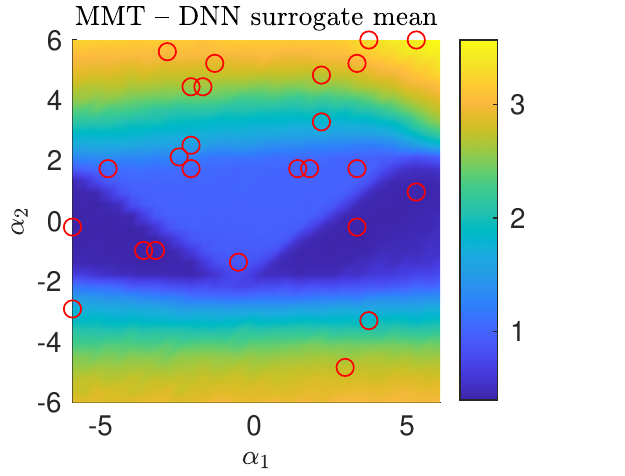}
\includegraphics[width=0.4\linewidth, trim=0 0 0 0, clip] {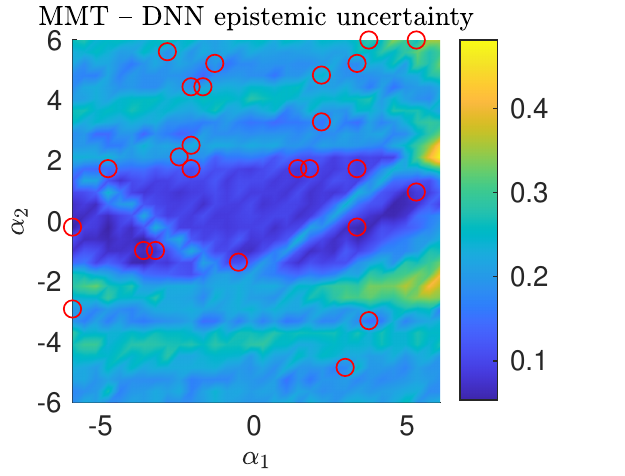}
\caption{Spatial distributions of the (\textbf{Left}) surrogate mean $\mu(\vec{\alpha})$ and (\textbf{Right}) epistemic uncertainty $\sigma_\epsilon(\vec{\alpha})$, for MMT data and different ML techniques.  $N=2, n_s=25$.}
\label{fig:mmt-comp-2d-surrogate}
\end{figure}

For completeness, in figure \ref{fig:mmt-comp-2d-surrogate} we present the corresponding epistemic plots for the MMT dataset for the GP, ENN and D-NN surrogate models.  The MMT system displays a simpler character than the LAMP system--in the 2D input restriction, the output is only weakly dependent on $\alpha_1$.

\bibliography{grad_bib.bib}
\bibliographystyle{plain}

\end{document}